\documentclass[10pt,twocolumn,letterpaper]{article}

\usepackage{cvpr}              % To produce the CAMERA-READY version
\usepackage{graphicx}
\usepackage{amsmath}
\usepackage{amssymb}
\usepackage{booktabs}
\usepackage{algorithm}
\usepackage{algpseudocode}
\usepackage{multirow}
\usepackage{placeins}

\usepackage{color}
\usepackage[dvipsnames]{xcolor}

\usepackage{xspace}
\usepackage{float}
\usepackage{contour}
\usepackage{xkeyval}
\usepackage{pifont}         % xmark, cmark

\usepackage[pagebackref,breaklinks,colorlinks]{hyperref}
\usepackage[abs]{overpic}

\usepackage[capitalize]{cleveref}
\crefname{section}{Sec.}{Secs.}
\Crefname{section}{Section}{Sections}
\Crefname{table}{Table}{Tables}
\crefname{table}{Tab.}{Tabs.}
\newcommand{\ouralg}{FvOR\xspace}

\contourlength{.4pt}
 \newcommand{\figlabel}[1]{\sffamily\bfseries\footnotesize\contour{white}{#1}} % label
\makeatletter
\newlength{\sfp@hseplen}\newlength{\sfp@vseplen}
\define@cmdkey{subfigpos}[sfp@]{vsep}[0.6\baselineskip]{\setlength{\sfp@vseplen}{\sfp@vsep}}% \sfp@vsep
\define@cmdkey{subfigpos}[sfp@]{hsep}[40pt]{\setlength{\sfp@hseplen}{\sfp@hsep}}% \sfp@hsep
\newcommand{\subfigimg}[3][,]{%
  \setkeys{Gin,subfigpos}{vsep,hsep,#1}% Set (default) keys
  \setbox1=\hbox{\includegraphics{#3}}% Store image in box
  \leavevmode\rlap{\usebox1}% Print image
  \rlap{\hspace*{\sfp@hsep}\raisebox{\dimexpr\ht1-80pt}{\figlabel{#2}}}% Print label
  \phantom{\usebox1}% Insert appropriate spacing
}
\makeatother

\def\cvprPaperID{5429} % *** Enter the CVPR Paper ID here
\def\confName{CVPR}
\def\confYear{2022}

\begin{document}

\title{HM3D-ABO: A Photo-realistic Dataset for Object-centric Multi-view 3D Reconstruction\\Technical Report}
\author{
  \hspace{-1.3cm}
  \begin{tabular}[t]{c}
    Zhenpei Yang$^{*}$\quad Zaiwei Zhang \quad Qixing Huang\\
    The University of Texas at Austin \\
    \end{tabular}
}
\maketitle
\let\thefootnote\relax\footnotetext{$^*$ yzp@utexas.edu}

\maketitle
\renewcommand{\baselinestretch}{0.9}

\begin{abstract}
Reconstructing 3D objects is an important computer vision task that has wide application in AR/VR. Deep learning algorithm developed for this task usually relies on an unrealistic synthetic dataset, such as ShapeNet~\cite{shapenet,choy20163d} and Things3D~\cite{xie2020pix2vox++}. On the other hand, existing real-captured object-centric datasets usually do not have enough annotation to enable supervised training or reliable evaluation. In this technical report, we present a photo-realistic object-centric dataset HM3D-ABO. It is constructed by composing realistic indoor scene~\cite{ramakrishnan2021habitat} and realistic object~\cite{collins2021abo}. For each configuration, we provide multi-view RGB observations, a water-tight mesh model for the object, ground truth depth map and object mask. The proposed dataset could also be useful for tasks such as camera pose estimation and novel-view synthesis. The dataset generation code is released at \url{https://github.com/zhenpeiyang/HM3D-ABO}.

\end{abstract}

\section{Introduction}
Reconstructing 3D object has been studied for decades~\cite{yang2019cubeslam, kundu20183d, lmvs,apple,peng2020neural,oechsle2021unisurf,wang2021neus,yang2021deep,yu2021pixelnerf,disn} and has been receiving more and more interest due to the recent popularity AR/VR applications. With multi-view renderings from object model collection datasets~\cite{shapenet,wu20153d,downs2022google,collins2021abo,gao2021objectfolder,Reizenstein_2021_ICCV}, researchers can develop data-driven approaches for object reconstruction but the clean backgrounds introduce a domain gap with real-world images. Things3D~\cite{xie2020pix2vox++} has tried to attach different backgrounds to those object renderings by placing CAD models in synthetic 3D indoor scenes. However, their renderings of synthetic indoor environments is still far from realistic. On the other hand, several works have also tried to create 3D object annotations from real 2D images, such as Pix3D~\cite{pix3d} and Scan2CAD~\cite{Avetisyan_2019_CVPR}. Yet these works either only provide a single-view image~\cite{pix3d} setting or does not provide an accurate 3D model~\cite{Avetisyan_2019_CVPR}. Redwood~\cite{Choi2016} contains multi-view 2D images and 3D depth scans but it only provides reconstructed 3D models for very few instances. Recently, Google Objectron~\cite{ahmadyan2021objectron} dataset has released a collection of short, object-centric video clips with 3D object-level annotations but it does not provide reconstructed 3D models. In light of the challenge of creating accurate annotations for real-world object scans, in this technical report, we propose an approach to generate a photo-realistic object-centric dataset based on existing high-quality scene and object models.\looseness=-1

\begin{figure}
\centering
\begin{overpic}[width=\linewidth]{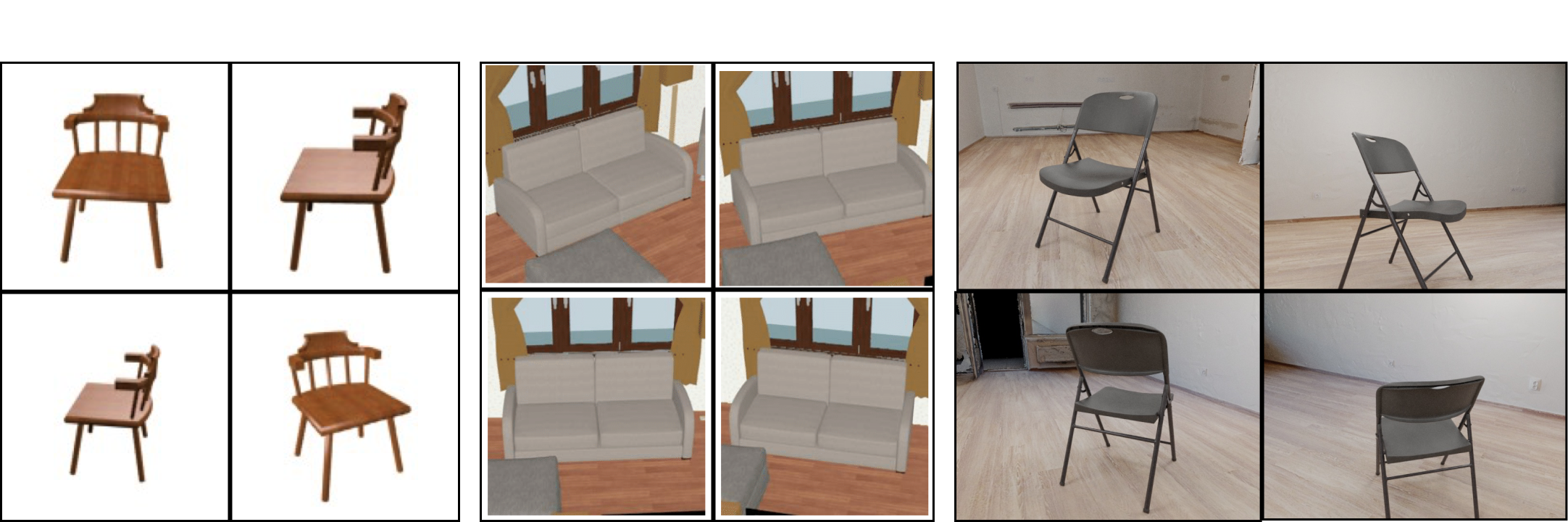}
\put(5,72){\footnotesize{ShapeNet Choy~\cite{choy20163d}}}
\put(84,72){\footnotesize{Things3D~\cite{xie2020pix2vox++}}}
\put(162,72){\footnotesize{HM3D-ABO}}
\vspace{-0.2in}
\end{overpic}
\vspace{-0.2in}
\caption{Qualitative comparison among three object-centric multi-view datasets. Our proposed HM3D-ABO dataset has both realistic foreground objects and realistic background scenes.  } 
\label{fig:dataset-compare}
\vspace{-0.1in}
\end{figure}
\section{HM3D-ABO Dataset}

\newcommand{\cmark}{\ding{51}}
\newcommand{\xmark}{\ding{55}}

\begin{table*}[ht!]
    \centering 
    
    \begingroup \setlength{\tabcolsep}{2pt}
    
    \begin{tabular}{l|c|ccc|cc}
        \toprule
        Method & Real & Multi-view  & Realistic Object &
        Realistic Scene& 
        Cam Pose & 3D Model 
        \\\hline
        Pix3D\cite{pix3d} & \cmark &\xmark & \cmark & \cmark & \cmark& \cmark \\
        
        Redwood\cite{Choi2016} &\cmark & \cmark & \cmark & \cmark & \xmark& \xmark \\
        
        Objectron\cite{ahmadyan2021objectron} &\cmark & \cmark & \cmark & \cmark & \cmark& \xmark \\
        
        ShapeNet Choy\cite{choy20163d} &\xmark & \cmark & \xmark & \xmark & \cmark& \cmark \\
        
        ShapeNet DISN\cite{choy20163d} &\xmark & \cmark & \xmark & \xmark & \cmark& \cmark \\
        
        Things3D\cite{xie2020pix2vox++}  &\xmark & \cmark & \xmark & \xmark & \cmark& \cmark \\\hline
        
        HM3D-ABO (Ours)  &\xmark & \cmark & \cmark & \cmark & \cmark & \cmark \\\hline
    \end{tabular}
    
    \endgroup
    \vspace{-0.1in}
    \caption{Comparison of the existing object-centric datasets with our proposed HM3D-ABO dataset. We compare based on whether it contains real or synthetic capture (Real), multi-view capture (Multi-view), realistic object (Realistic Object), realistic background scene (Realistic Scene), camera pose annotation (Cam Pose), and accurate 3D model for the object (3D Model). }
    \label{tab:nuscenes}
\end{table*}

\noindent\textbf{Dataset Assets.}
In order to build our dataset, we utilize two types of assets: scene and object. We use the recently released Habitat-Matterport 3D dataset~\cite{ramakrishnan2021habitat} as our scene assets. Those scenes are high quality capturing of real-world indoor scenes. We use Amazon-Berkeley Object~\cite{collins2021abo} dataset as our object assets, because it comes with high-quality texture and accurate geometry. Since their provided meshes are not all watertight and may contain some internal structures, we follow the procedure of OccNet~\cite{mescheder2019occupancy} to render hundreds of depth maps covering the object surface, then running TSDF-Fusion and marching cube~\cite{lorensen1987marching} to get the water-tight mesh. \looseness=-1

\noindent\textbf{Pipeline.}  The overall procedure of creating synthetic scans goes as follows. We randomly choose a scene and an object. Then we find a placement location of such an object in that scene by sampling the navigable position~\cite{szot2021habitat,habitat19iccv} in the scene. We avoid physical collision between the inserted object and the rest of the scene by ensuring there is an empty region in the surrounding of the chosen location. We also make sure that the object is placed on the floor of the scene instead of hovering in the air by running a few steps of physical simulation. We then sample a set of camera pose around the inserted object and ensure that the camera does not collide with the scene. We then render the image at each location using habitat-sim~\cite{szot2021habitat,habitat19iccv}'s built-in renderer. After rendering, we do a post-filtering to filter out low-quality capture. We calculate the ratio of the area of the bounding rectangle of the object mask to the size of the image and filter out images that have a ratio less than $0.2$. Finally, we use the physical-based rendering engine Blender Cycles~\cite{blender} to render the filtered configurations at $640\times 480$ resolution. Each image takes around 10 seconds to render using one NVIDIA V100 GPU.

\noindent\textbf{Statistics.} 
 HM3D-ABO has 1966 objects placed in 500 indoor scenes.  In total, there are 3196 scene-shape configurations. In Fig.~\ref{fig:dataset-compare} we show qualitative comparisons between HM3D-ABO and existing object-centric multi-view datasets. More examples of our dataset are shown in Fig.~\ref{fig:visualization:matterportabo}.

\section{Benchmark}
We benchmark two tasks with the HM3D-ABO dataset: absolute camera pose estimation and few-view 3D object reconstruction. Our experimental setup follows FvOR~\cite{yang2022fvor}.

\subsection{Camera Pose Estimation}
Camera pose estimation is a core step of many multi-view 3D reconstruction algorithm. Camera pose estimation approaches can be categorized into two types, i.e. relative pose approach and absolute pose approach. Relative pose approach estimates the relative pose between an image pair. Traditional approach relies on matching hand-crafted key-points~\cite{harris1988combined,lowe2004distinctive,bay2006surf}, which could not handle much appearance variation and requires large overlap between image pair. The advance of deep learning also leads to algorithms that reduce the sensitivity to appearance change and low visual overlap~\cite{sarlin2020superglue, sun2021loftr, yang2019extreme, yang2020extreme, cai2021extreme, jin2021planar, shabani2021extreme, Ma_2022_CVPR,sun2022global}. The second type is more tight to object-centric scenario. The absolute camera pose approach directly estimates the camera pose w.r.t. the canonical object coordinate system~\cite{disn,yang2022fvor}. Compared to relative pose estimation, these type of methods may be less generalizable compared to relative pose approach due to the assumption of canonical object coordinate system. But they are usually simpler as they directly output the camera poses for each image, without the need to perform pairwise relative camera pose estimation followed by synchronization step to get absolute camera pose. 

Here we provide a benchmark of absolute camera pose estimation. The input is a bag of RGB images capturing an object, and the output is the absolute camera pose for each of the images, w.r.t the canonical object coordinate. We follow the experimental setup of FvOR~\cite{yang2022fvor, disn} and use 4 images as input. We report the mean and median pixel error, rotation error, and translation error for each algorithm. The pixel error is computed w.r.t a 512x384 input image. We benchmark three methods as in~\cite{yang2022fvor}. The first one is based on DISN~\cite{disn} which parametrizes camera poses by orthogonal vectors. The second one is based on Cai et.al~\cite{cai2021extreme} which use a classification plus regression approach. The last one we run is from FvOR~\cite{yang2022fvor}'s pose initialization module(denoted as FvOR-Pose in the following text), which uses a scene-coordinate as intermediate pose representation followed by RANSAC PnP~\cite{opencv_library} for pose extraction. The results can be found in Table~\ref{table:habitat-pose}.   \looseness=-1

\begin{table}
\centering
\resizebox{0.95\linewidth}{!}{%
\begin{tabular}{rcccc}
\toprule

\multicolumn{1}{c}{Method}       & Pixel Error$\downarrow$ & Rot Error($^\circ$)$\downarrow$ & Trans Error(cm)$\downarrow$\\ \hline

DISN~\cite{disn} &  23.0/9.6  & 18.6/4.1 & 9.1/6.2 \\

Cai~\etal~\cite{cai2021extreme} &38.9/18.0  & 31.3/7.7 & 12.3/9.9 \\

\ouralg-Pose~\cite{yang2022fvor} &  \textbf{18.0}/\textbf{5.0}  & \textbf{14.3}/\textbf{1.3}  &\textbf{7.6}/\textbf{5.0} \\%
\bottomrule
\end{tabular}
}
\vspace{-0.1in}
\caption{Absolute camera pose estimation results on HM3D-ABO dataset. We show the mean/median statistics for all metrics. }
\label{table:habitat-pose}
\vspace{-0.1in}
\end{table}

\begin{table*}
\centering
\resizebox{0.99\textwidth}{!}{%
\begin{tabular}{r|cc|cc cc cc|cc}
\toprule
\multirow{2}{*}{Method}   & \multicolumn{2}{c|}{GT}  & \multicolumn{2}{c}{Noise@L1}  & \multicolumn{2}{c}{Noise@L2} & \multicolumn{2}{c|}{Noise@L3} & \multicolumn{2}{c}{Predict}  \\ 

   & \multicolumn{1}{c}{IoU$\uparrow$} & Chamfer-L1$\downarrow$  & \multicolumn{1}{c}{IoU$\uparrow$} & Chamfer-L1$\downarrow$  & \multicolumn{1}{c}{IoU$\uparrow$} & Chamfer-L1$\downarrow$  & \multicolumn{1}{c}{IoU$\uparrow$} & Chamfer-L1$\downarrow$ & \multicolumn{1}{c}{IoU$\uparrow$} & Chamfer-L1$\downarrow$\\ \hline

OccNet$^\dagger$~\cite{mescheder2019occupancy,apple}    &0.762/0.812 &2.01/1.63& 0.762/0.812 &2.01/1.63&0.762/0.812 &2.01/1.63& 0.762/0.812 &2.01/1.63& 0.762/0.812 &2.01/1.63 \\

IDR~\cite{idr}    & 0.645/0.677 &3.67/2.78&0.643/0.684 &3.75/2.90&0.642/0.679& 3.70/2.84 & 0.638/0.665& 3.78/3.00& 0.612/0.650& 4.40/3.37 \\

\ouralg w/ GT Pose~\cite{yang2022fvor}   & \textbf{0.861}/\textbf{0.891} &\textbf{0.791}/\textbf{0.680}& 0.821/0.863 &1.22/1.05& 0.761/0.815& 2.04/1.76 & 0.703/0.768 &2.90/2.47 & 0.816/0.870& 1.40/0.93\\

\ouralg w/o Joint~\cite{yang2022fvor}   & \underline{0.859}/\underline{0.889}   &\underline{0.855}/\underline{0.741}&\underline{0.849}/\underline{0.882} &\underline{0.944}/\underline{0.825}& \underline{0.819}/\underline{0.864}& \underline{1.28}/\underline{1.16} & \underline{0.776}/\underline{0.831} & \underline{1.90}/1.65 & \underline{0.834}/\underline{0.877}& \underline{1.21}/\underline{0.886} \\

\ouralg~\cite{yang2022fvor}   & 0.853/0.885&0.921/0.793& \textbf{0.852}/\textbf{0.883} &\textbf{0.941}/\textbf{0.806}& \textbf{0.849}/\textbf{0.880}& \textbf{0.980}/\textbf{0.843} & \textbf{0.845}/\textbf{0.876} &\textbf{1.05}/\textbf{0.897} & \textbf{0.839}/\textbf{0.878}&\textbf{1.19}/\textbf{0.867}  \\
\bottomrule
\end{tabular}
}
\vspace{-0.1in}

\caption{Evaluating the robustness of few-view 3D reconstruction on HM3D-ABO dataset (mean/median, top-2 results highlighted). We report the results using ground truth poses, perturbed poses with different perturbation levels, and predicted poses by FvOR-Pose. Chamfer-L1 is multiplied by 100. }
\label{table:habitat-shape-refine}
\vspace{-0.1in}
\end{table*}

\begingroup
\setlength{\tabcolsep}{1.0pt} % Default value: 6pt
   \begin{figure*}
    %  \centering
    \hspace{-1em}
     \def\imw{0.14\textwidth}
     \def\imh{0.08\textheight}

     \begin{tabular}{ccccccc}

        \includegraphics[height=\imh, width=\imw, keepaspectratio]{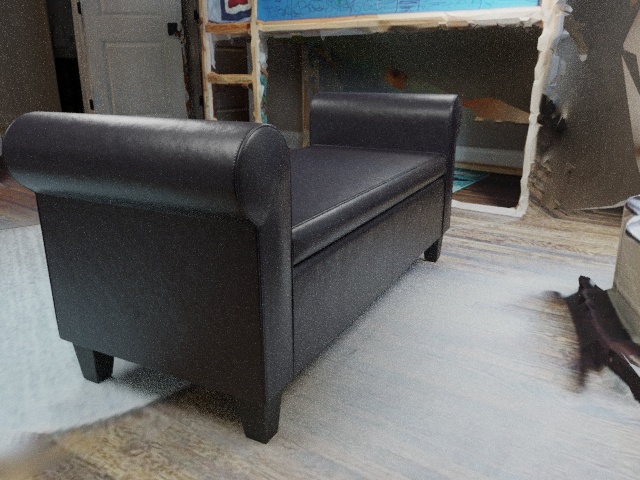} & 
     \includegraphics[height=\imh, width=\imw, keepaspectratio]{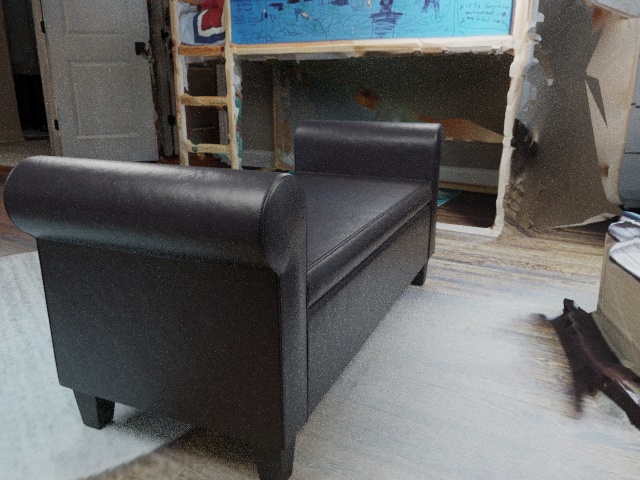} &
     \includegraphics[height=\imh, width=\imw, keepaspectratio]{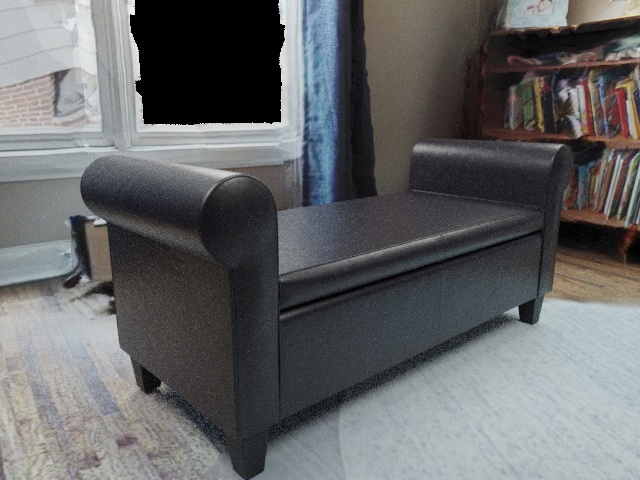} &
     \includegraphics[height=\imh, width=\imw, keepaspectratio]{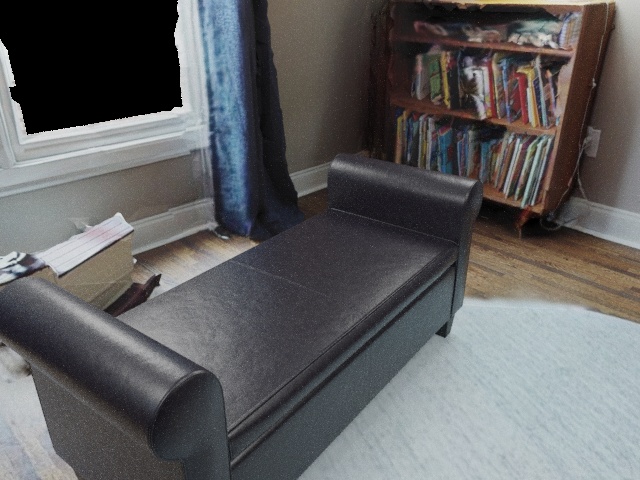} &
      \includegraphics[height=\imh, width=\imw, keepaspectratio]{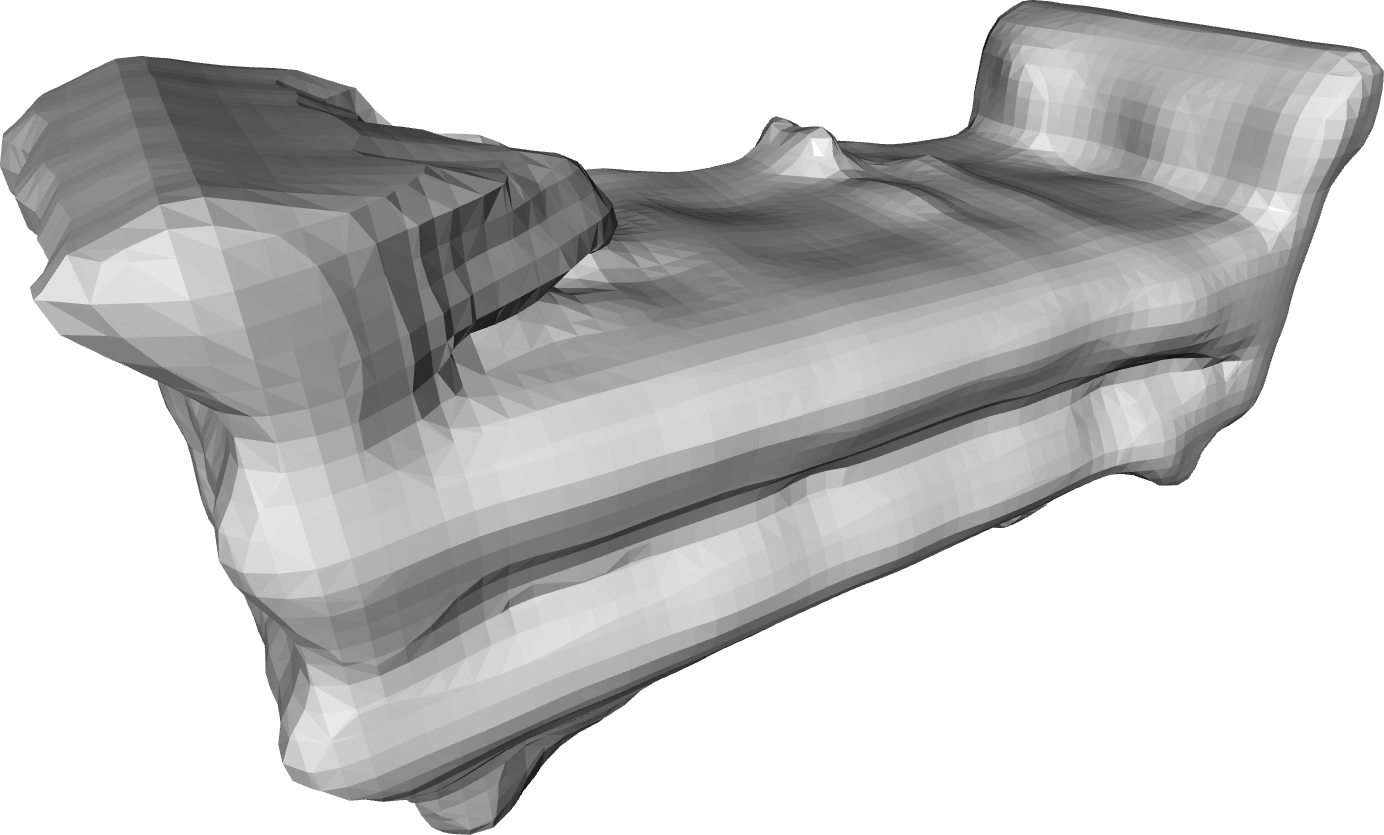}&
        \includegraphics[height=\imh, width=\imw, keepaspectratio]{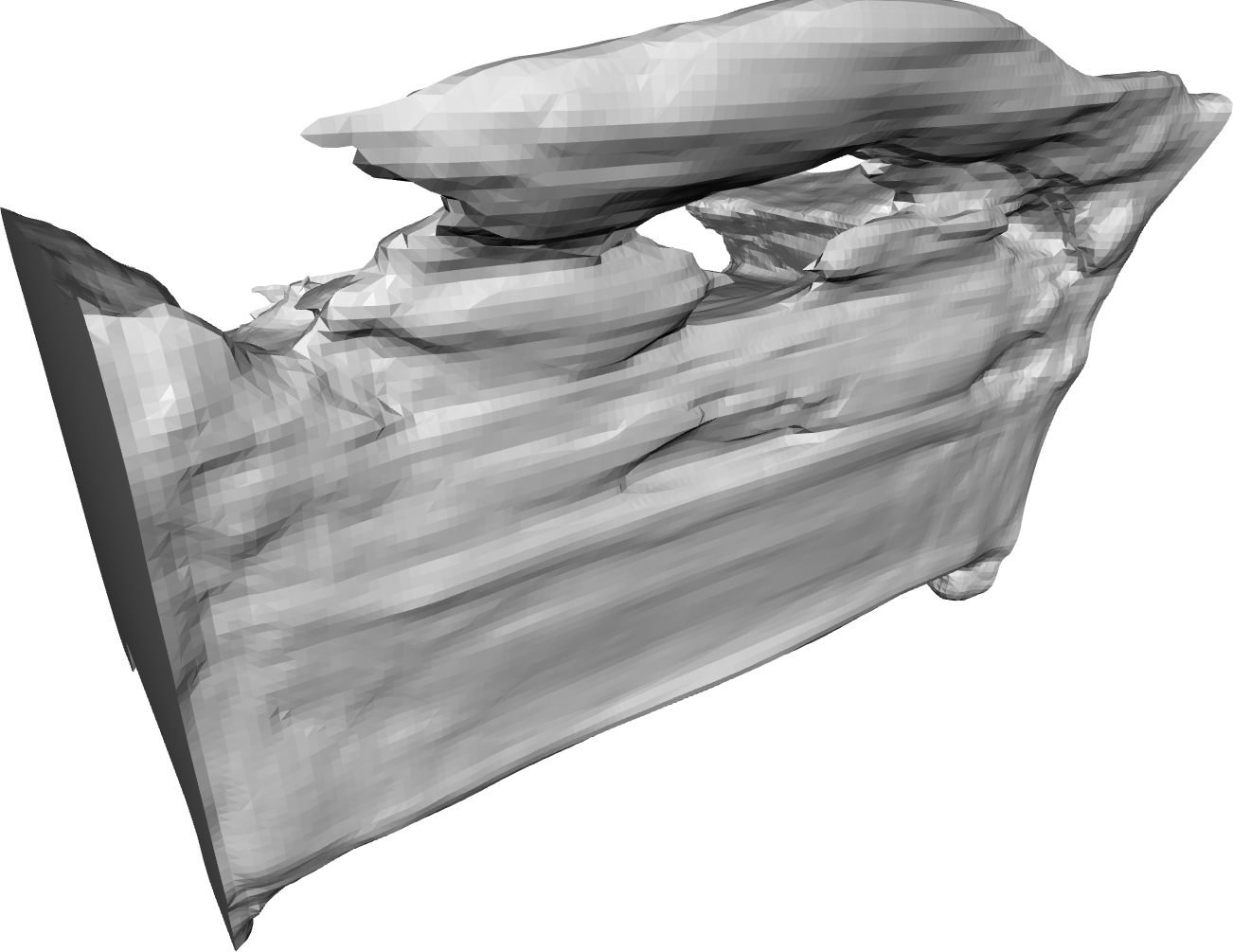}&
        \includegraphics[height=\imh, width=\imw, keepaspectratio]{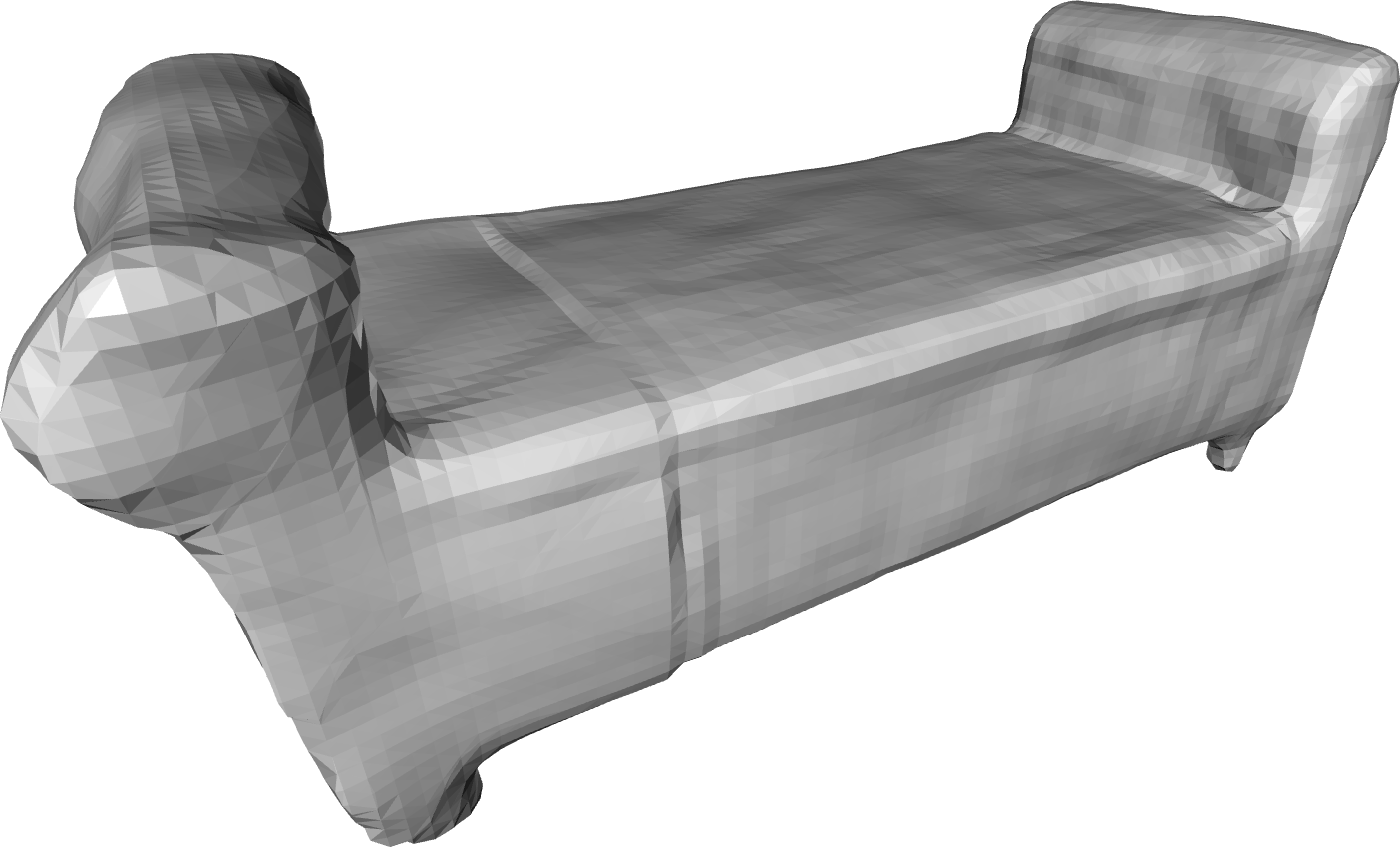} \\

     \includegraphics[height=\imh, width=\imw, keepaspectratio]{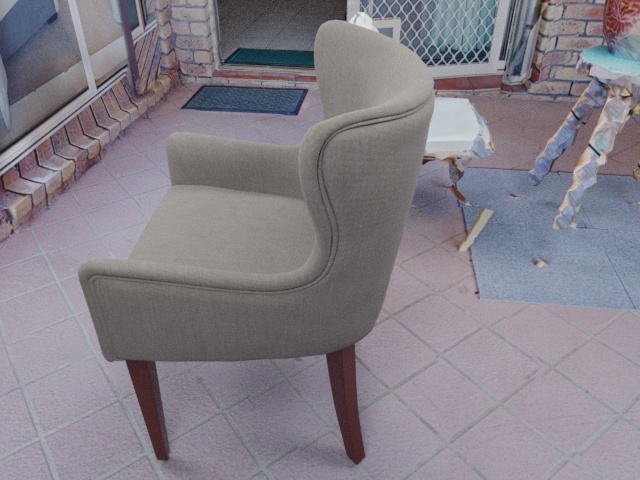} & 
     \includegraphics[height=\imh, width=\imw, keepaspectratio]{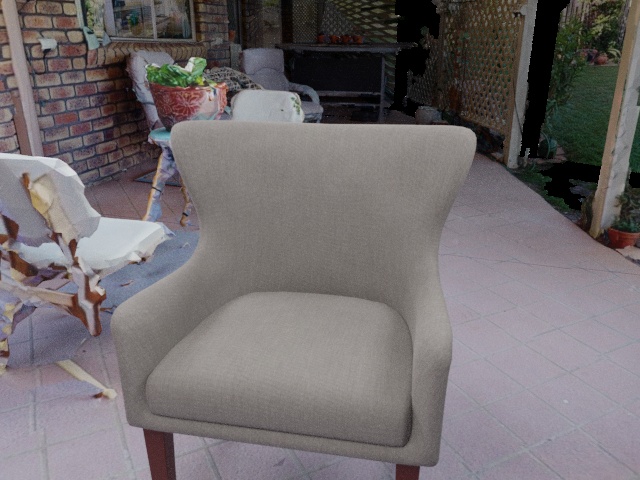} &
     \includegraphics[height=\imh, width=\imw, keepaspectratio]{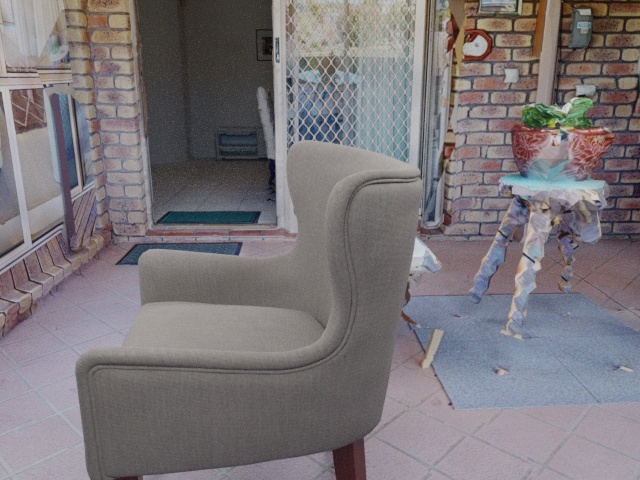} &
     \includegraphics[height=\imh, width=\imw, keepaspectratio]{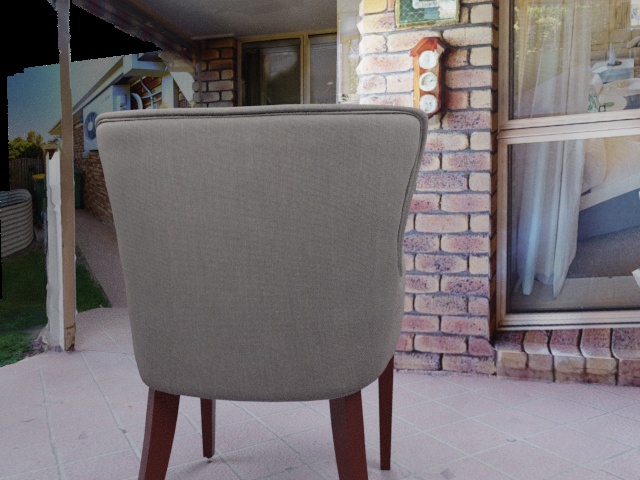} &
      \includegraphics[height=\imh, width=\imw, keepaspectratio]{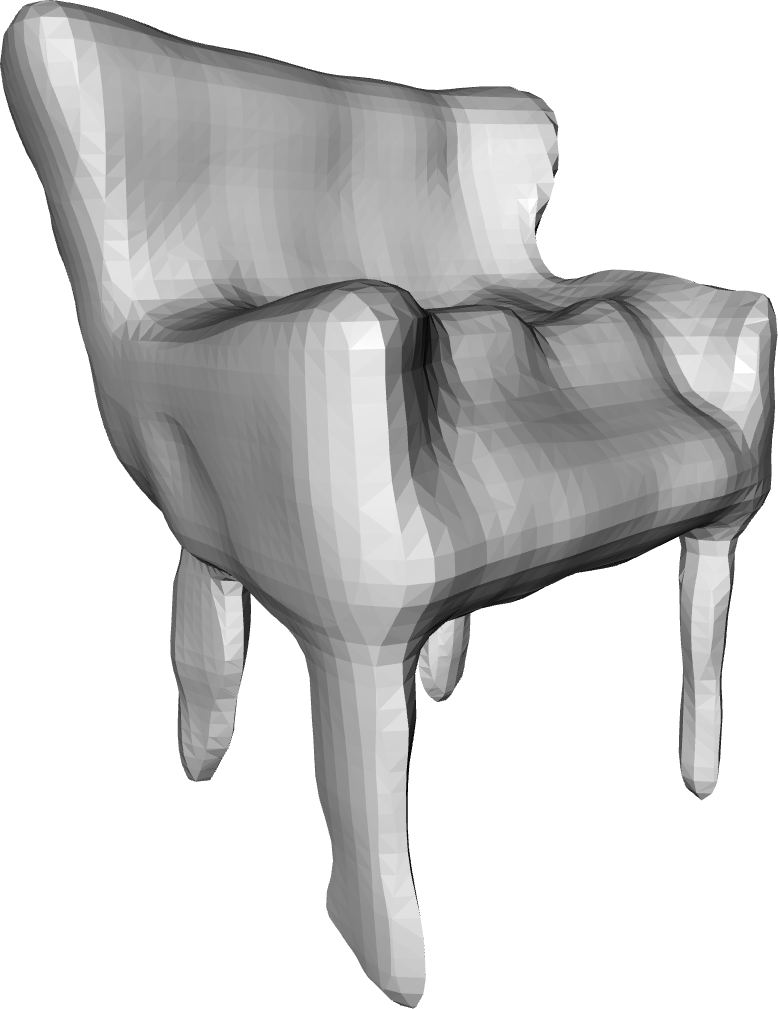}&
        \includegraphics[height=\imh, width=\imw, keepaspectratio]{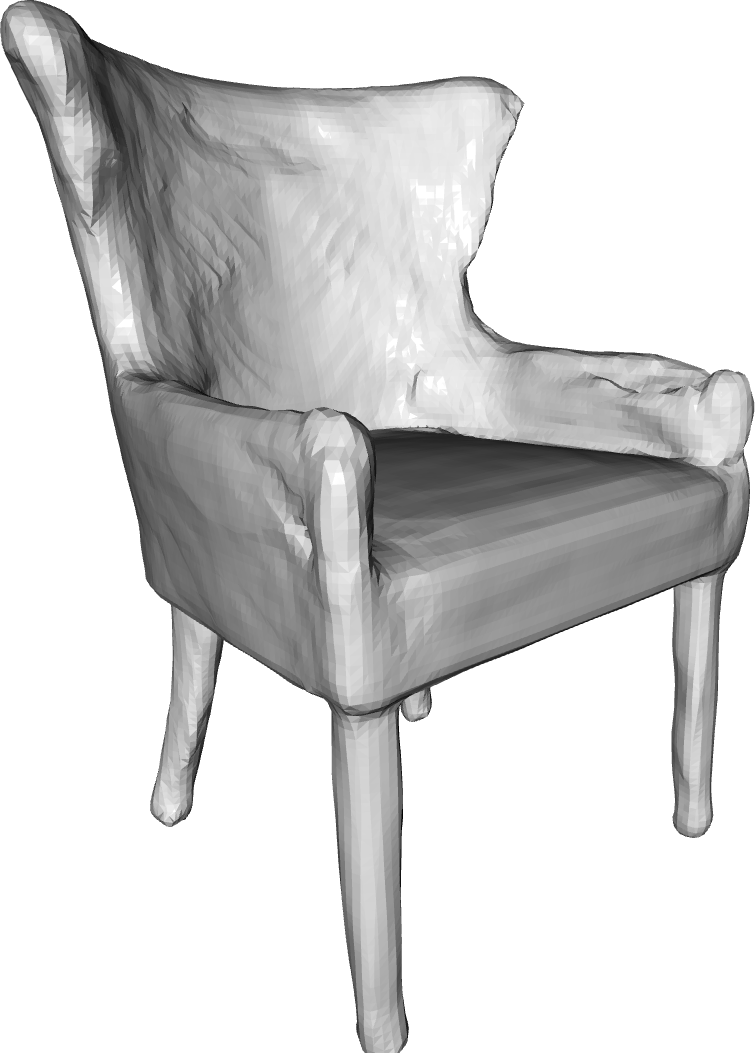}&
        \includegraphics[height=\imh, width=\imw, keepaspectratio]{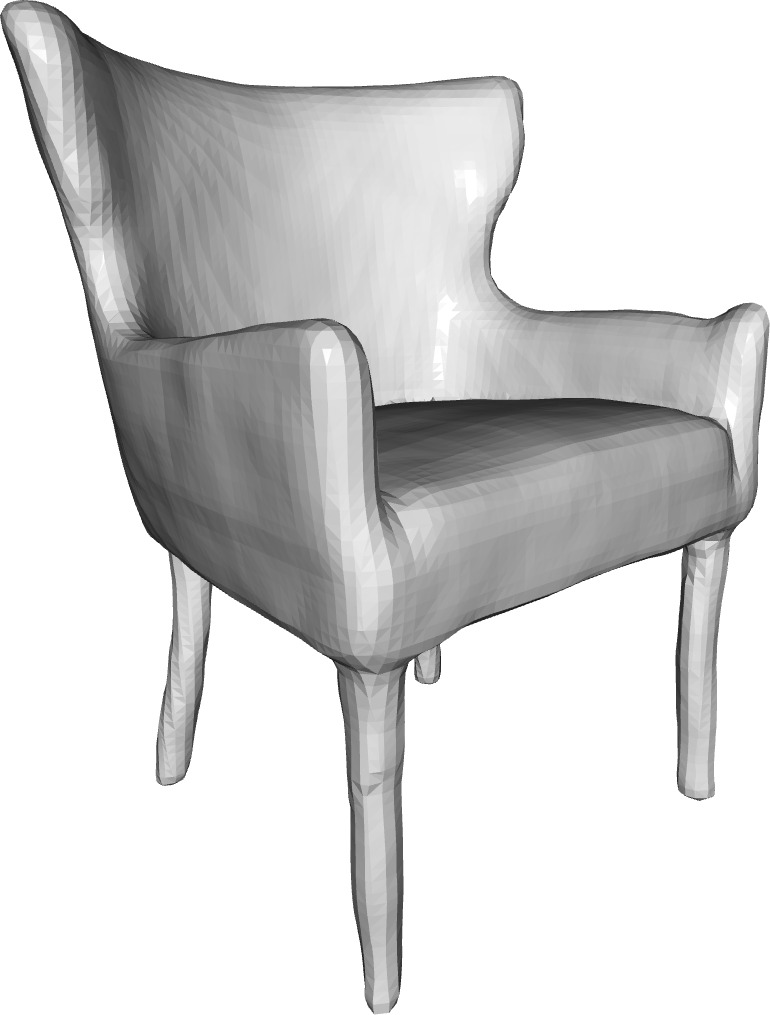} \\

     \includegraphics[height=\imh, width=\imw, keepaspectratio]{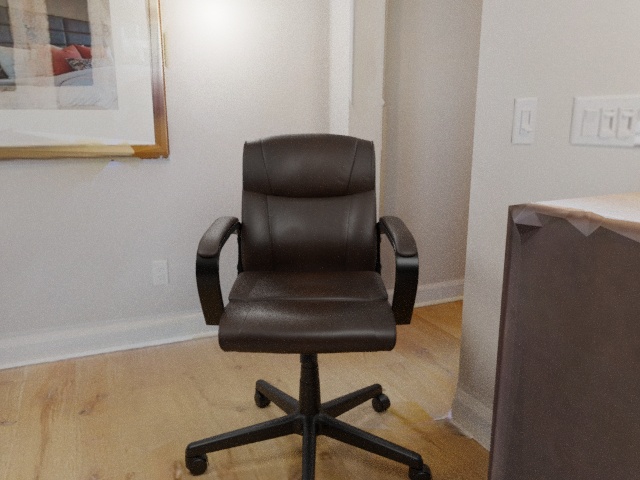} & 
     \includegraphics[height=\imh, width=\imw, keepaspectratio]{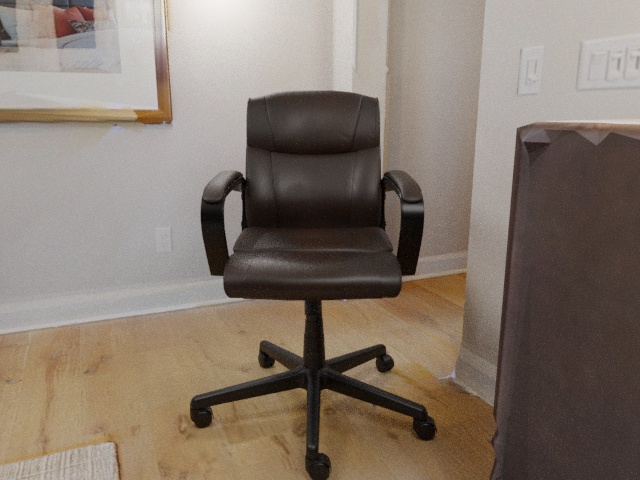} &
     \includegraphics[height=\imh, width=\imw, keepaspectratio]{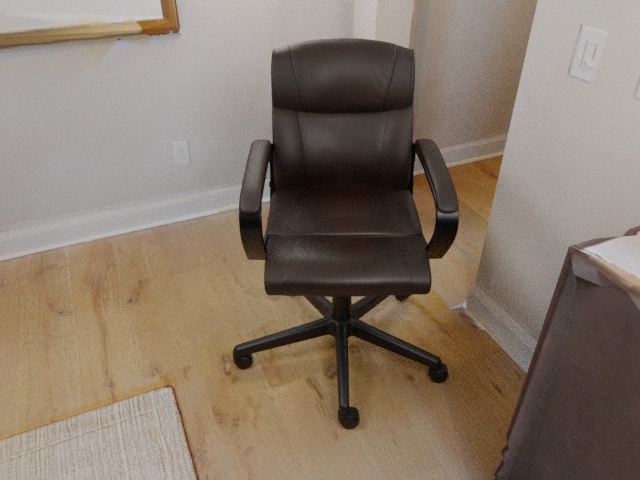} &
     \includegraphics[height=\imh, width=\imw, keepaspectratio]{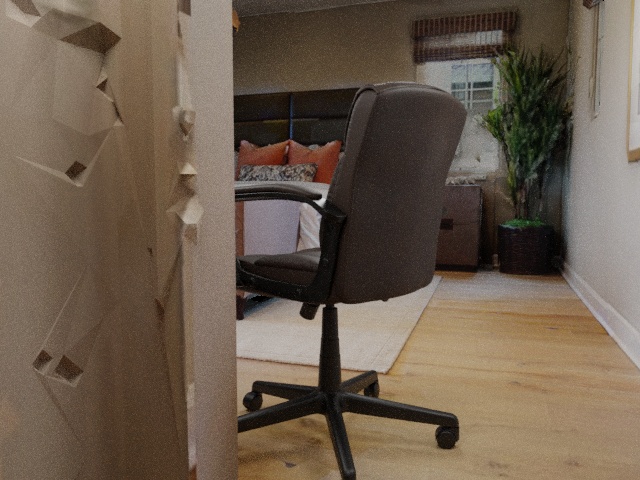} &
      \includegraphics[height=\imh, width=\imw, keepaspectratio]{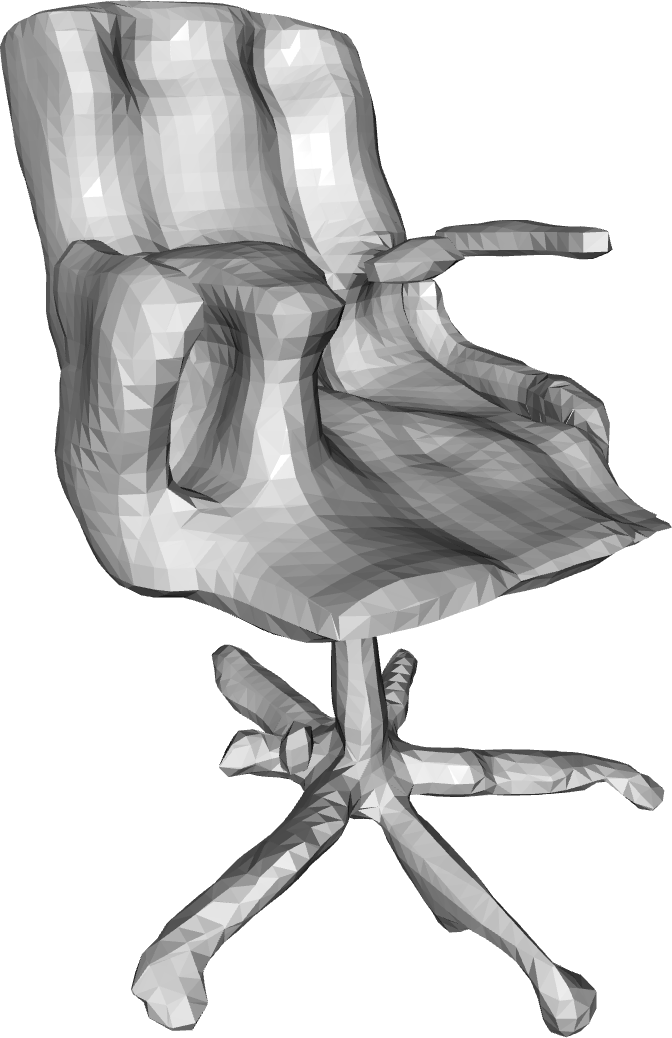}&
        \includegraphics[height=\imh, width=\imw, keepaspectratio]{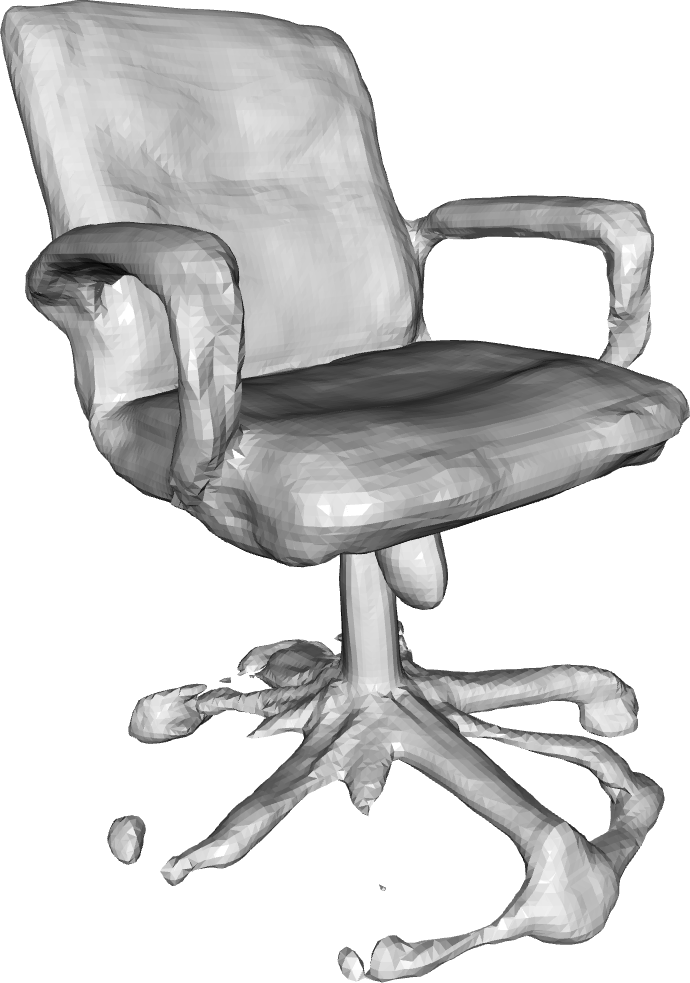}&
        \includegraphics[height=\imh, width=\imw, keepaspectratio]{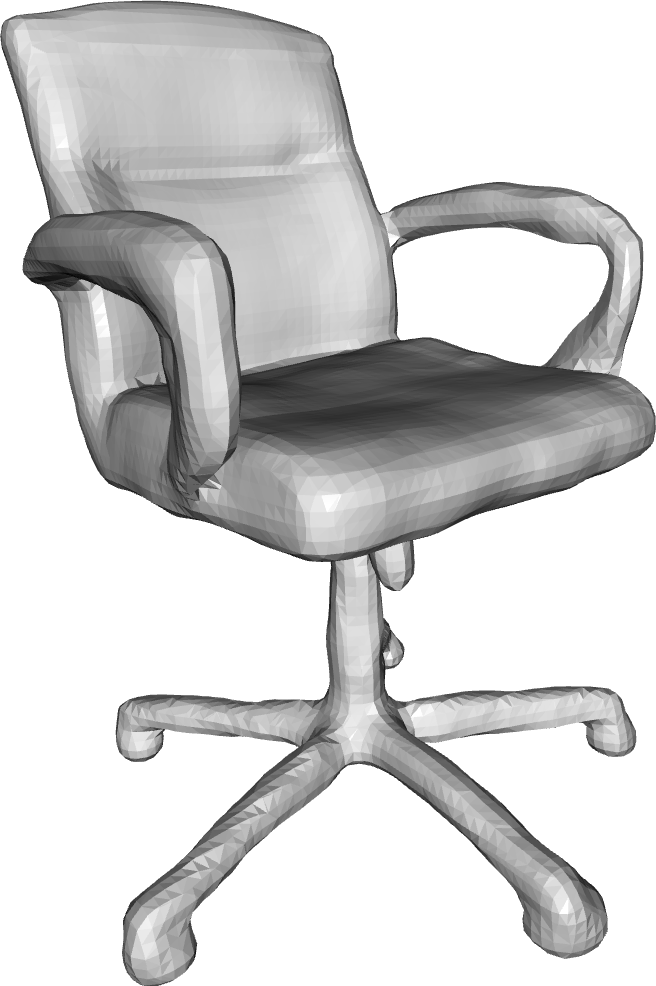} \\

     \includegraphics[height=\imh, width=\imw, keepaspectratio]{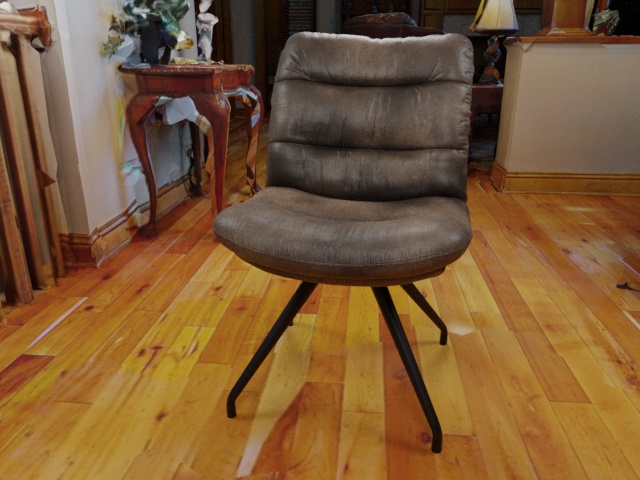} & 
     \includegraphics[height=\imh, width=\imw, keepaspectratio]{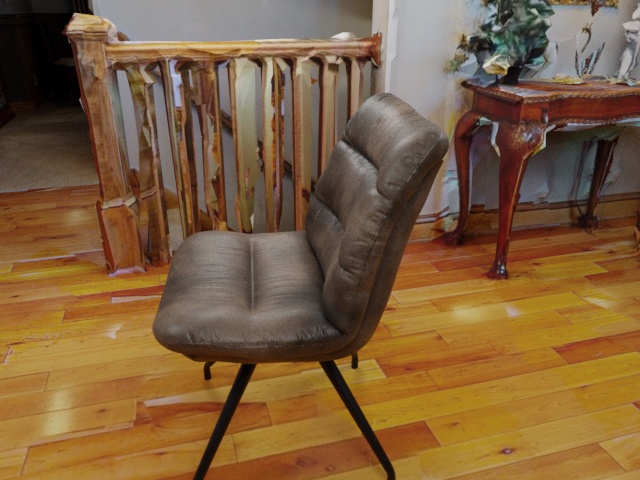} &
     \includegraphics[height=\imh, width=\imw, keepaspectratio]{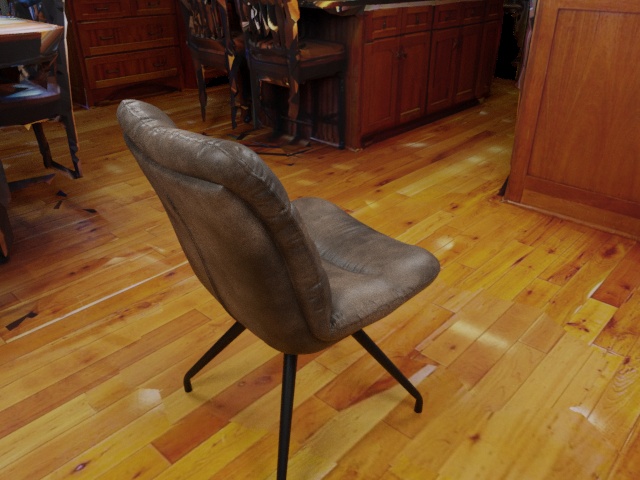} &
     \includegraphics[height=\imh, width=\imw, keepaspectratio]{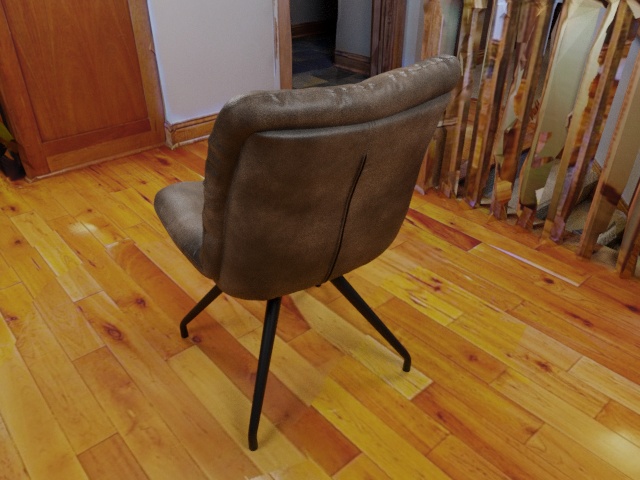} &
      \includegraphics[height=\imh, width=\imw, keepaspectratio]{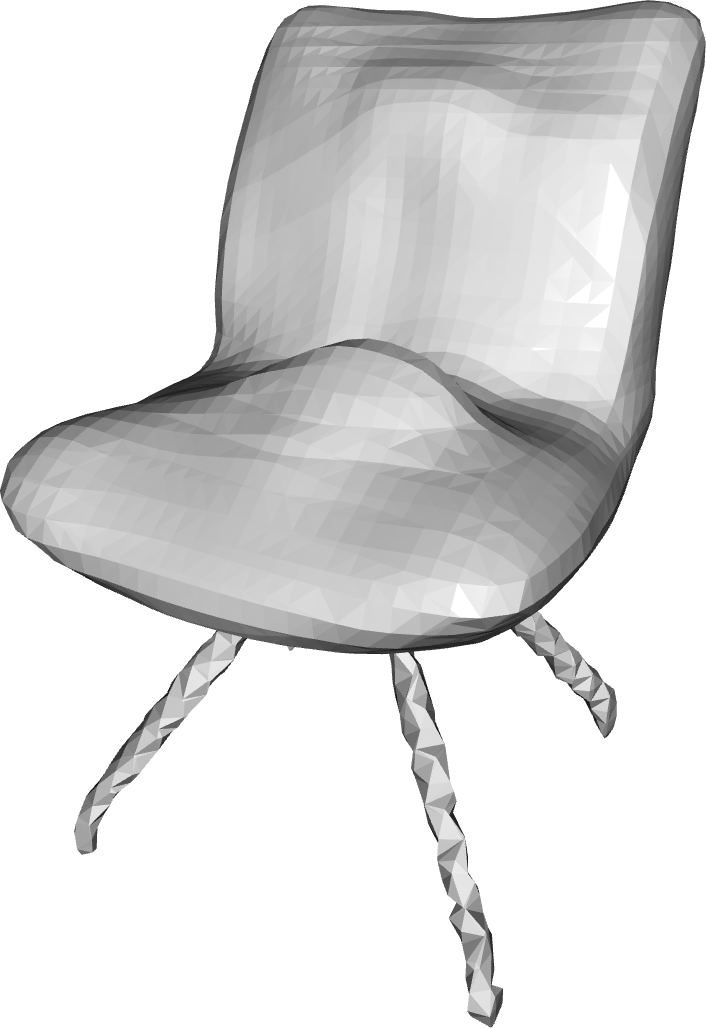}&
        \includegraphics[height=\imh, width=\imw, keepaspectratio]{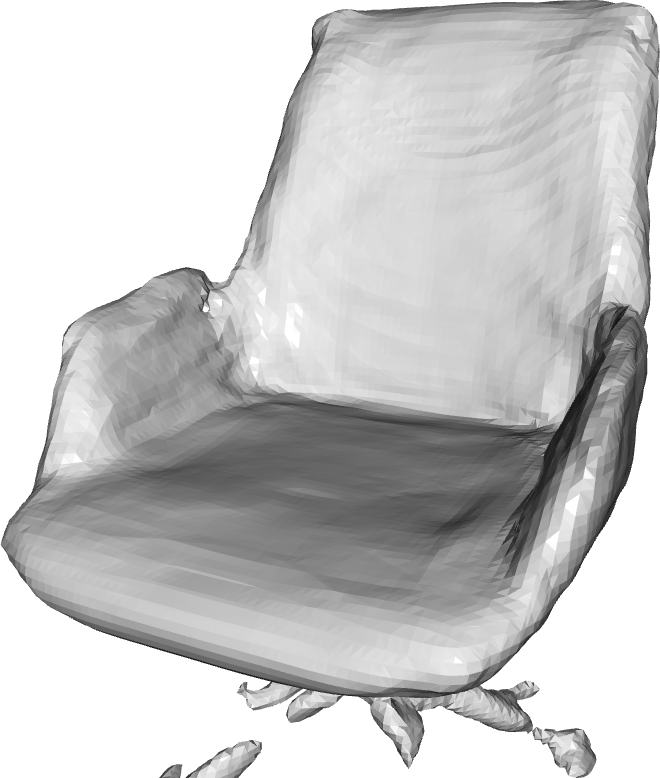}&
        \includegraphics[height=\imh, width=\imw, keepaspectratio]{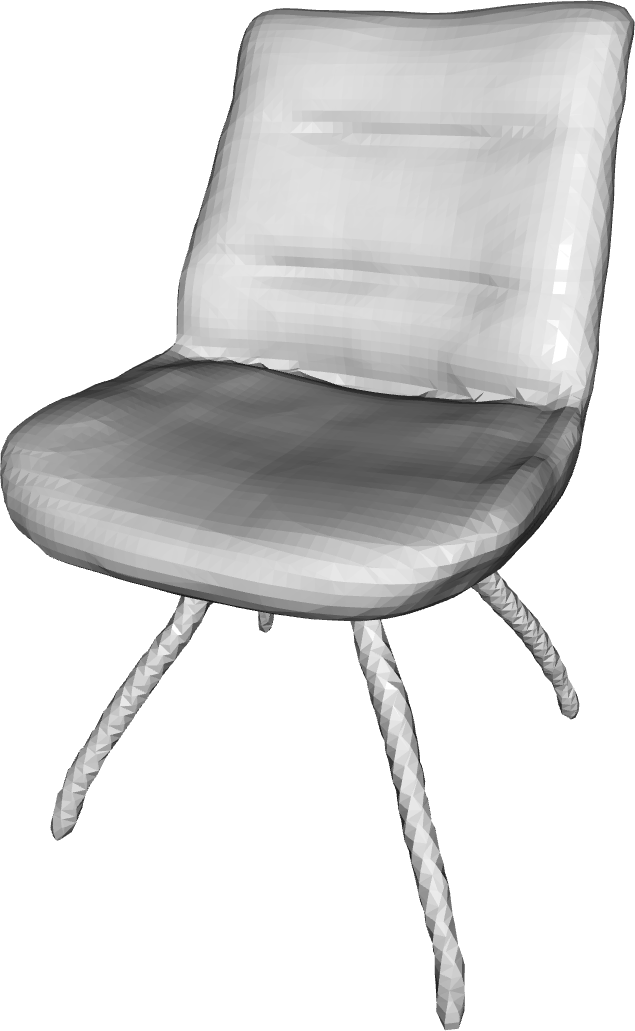} \\
        
     \includegraphics[height=\imh, width=\imw, keepaspectratio]{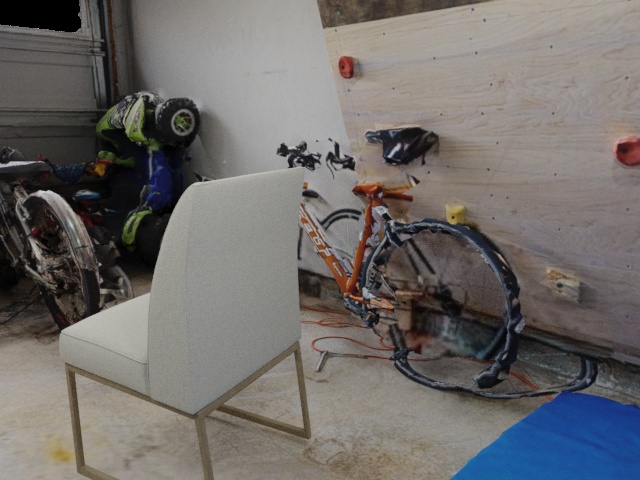} & 
     \includegraphics[height=\imh, width=\imw, keepaspectratio]{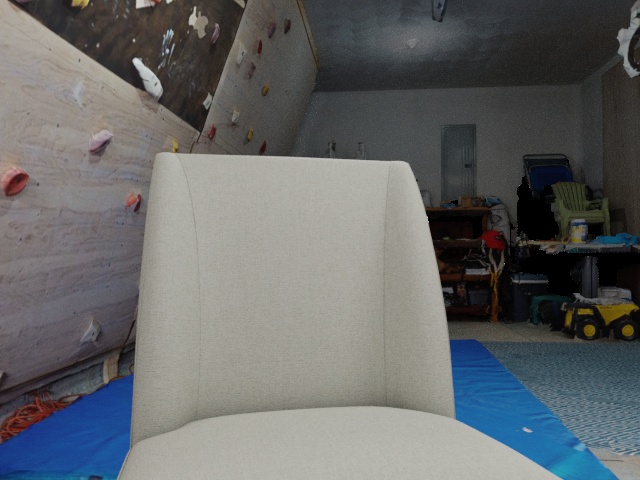} &
     \includegraphics[height=\imh, width=\imw, keepaspectratio]{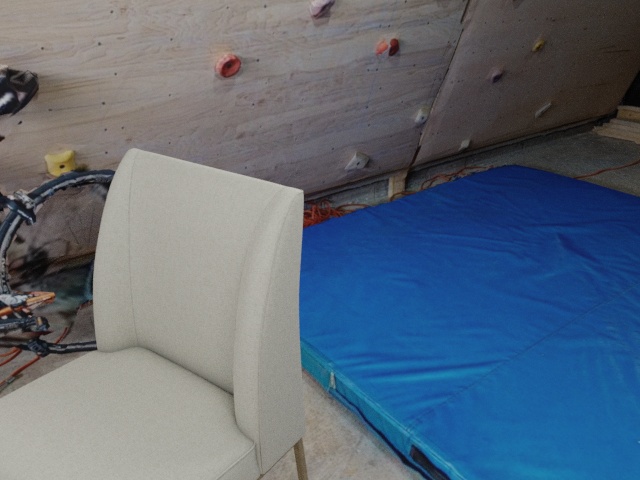} &
     \includegraphics[height=\imh, width=\imw, keepaspectratio]{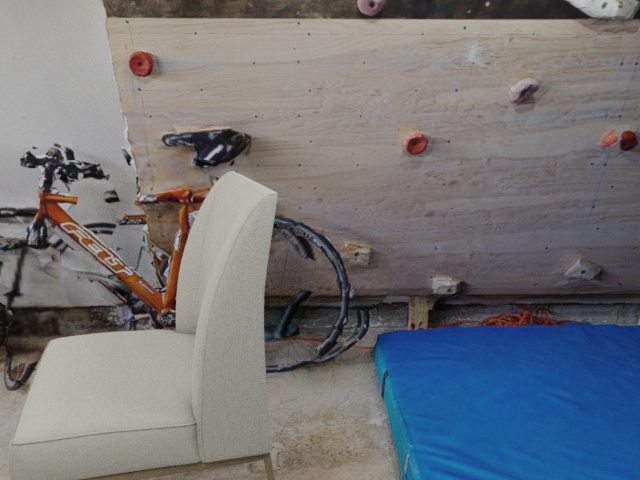} &
      \includegraphics[height=\imh, width=\imw, keepaspectratio]{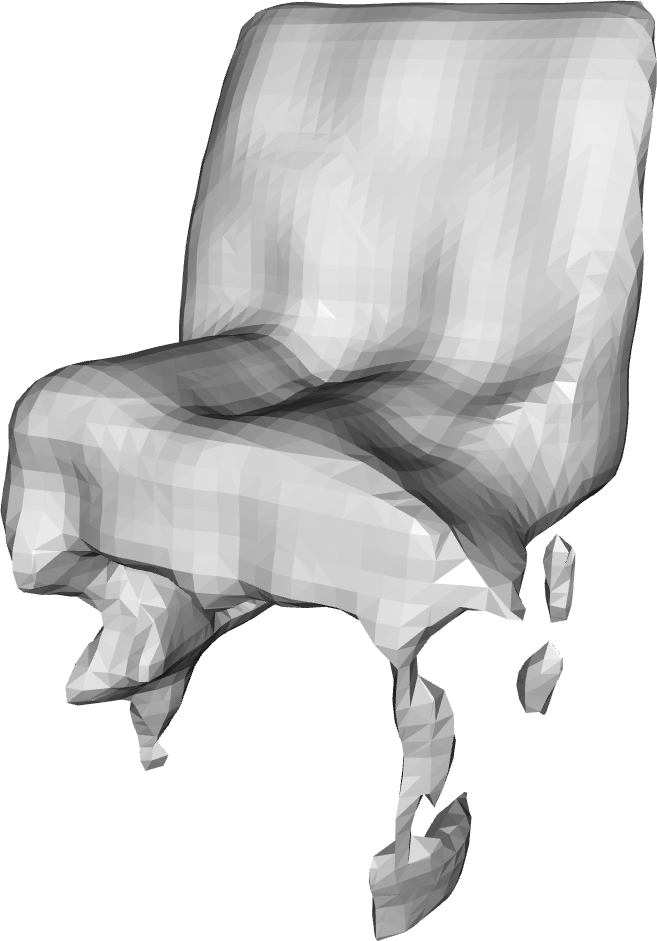}&
        \includegraphics[height=\imh, width=\imw, keepaspectratio]{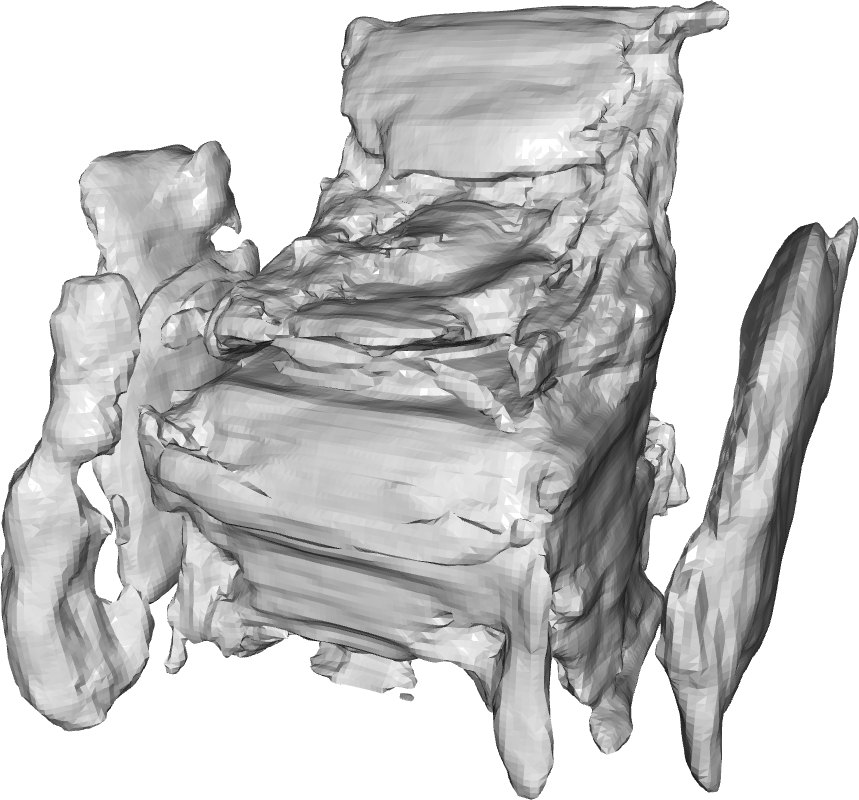}&
        \includegraphics[height=\imh, width=\imw, keepaspectratio]{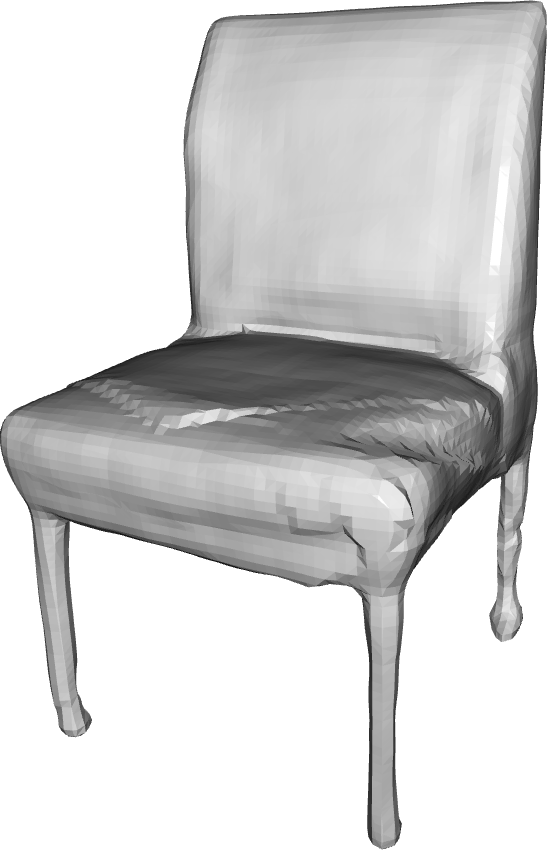} \\

             \multicolumn{4}{c}{Few-view Input} & IDR~\cite{idr} & OccNet$^\dagger$~\cite{mescheder2019occupancy,apple} &FvOR~\cite{yang2022fvor}
	 \end{tabular}
     \vspace{-5pt}
     \caption{Qualitative comparisons between three approaches on HM3D-ABO. For each column, we show the 4 input views on the left and the reconstruction of each method on the right. FvOR~\cite{yang2022fvor} and IDR~\cite{idr} use pose predicted by FvOR-Pose~\cite{yang2022fvor}.}
     \label{fig:visualization}\vspace{-1mm}
   \end{figure*}
\endgroup

\begingroup
\setlength{\tabcolsep}{1.0pt} % Default value: 6pt
   \begin{figure*}[!ht]
    \centering
    \hspace{-1em}
     \def\imw{0.28\textwidth}
     \def\imh{0.13\textheight}

     \begin{tabular}{cccc}
        \includegraphics[height=\imh, width=\imw, keepaspectratio]{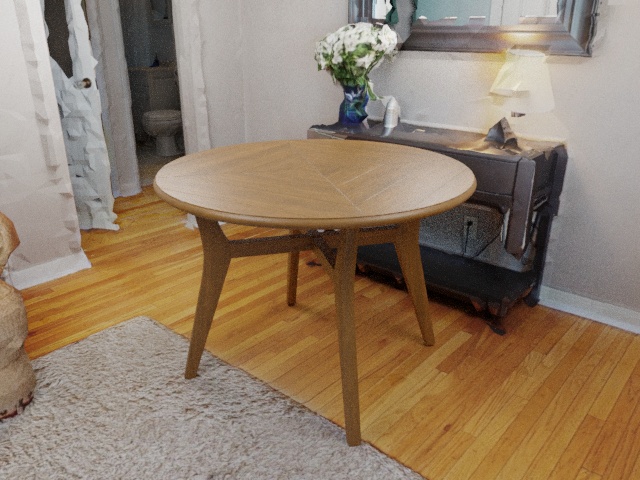} & 
         \includegraphics[height=\imh, width=\imw, keepaspectratio]{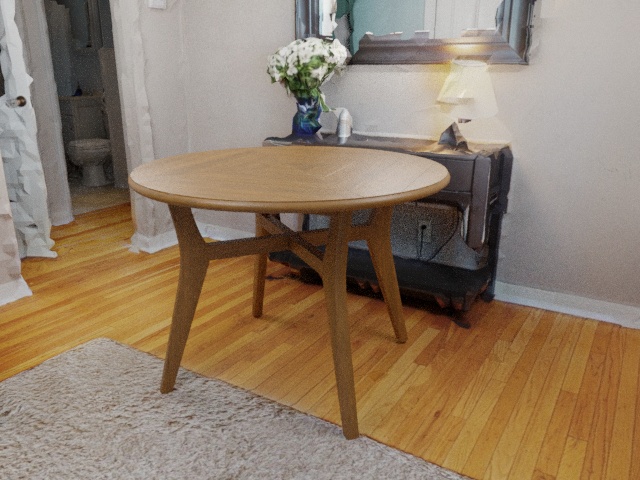} &
          \includegraphics[height=\imh, width=\imw, keepaspectratio]{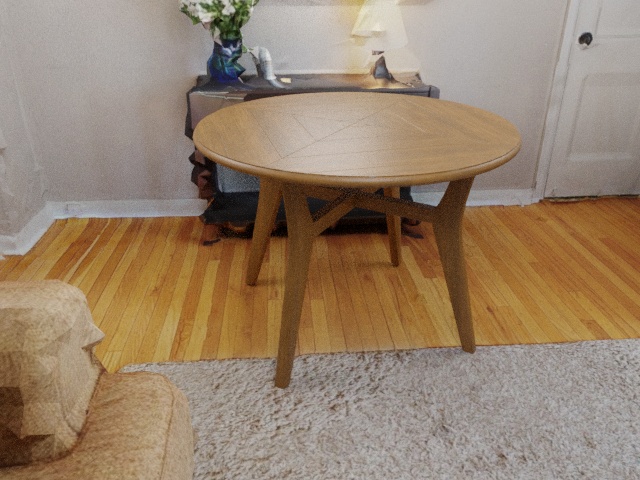} &
           \includegraphics[height=\imh, width=\imw, keepaspectratio]{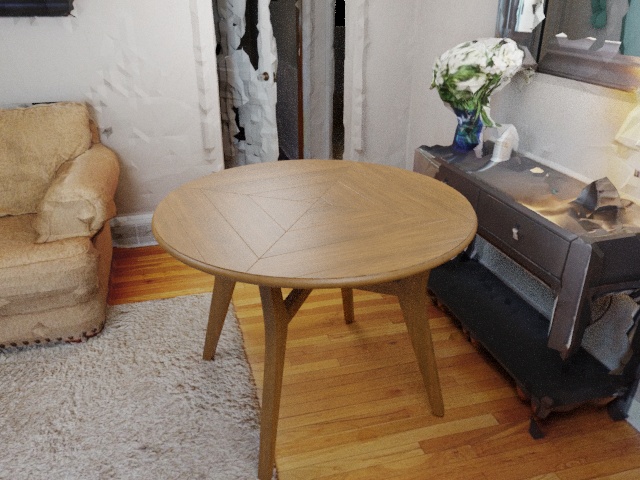} \\
           
           \includegraphics[height=\imh, width=\imw, keepaspectratio]{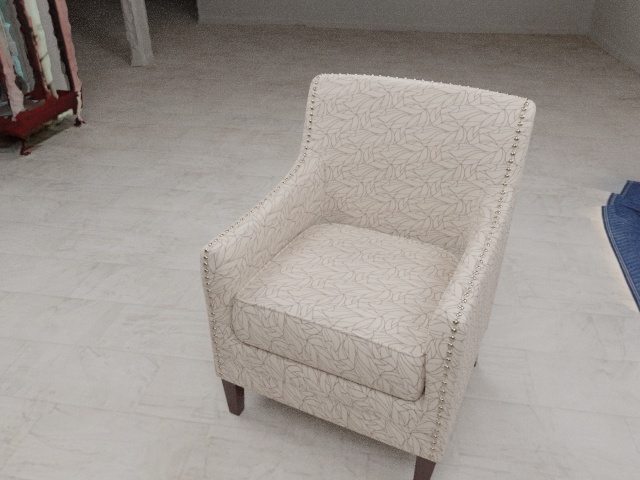} & 
         \includegraphics[height=\imh, width=\imw, keepaspectratio]{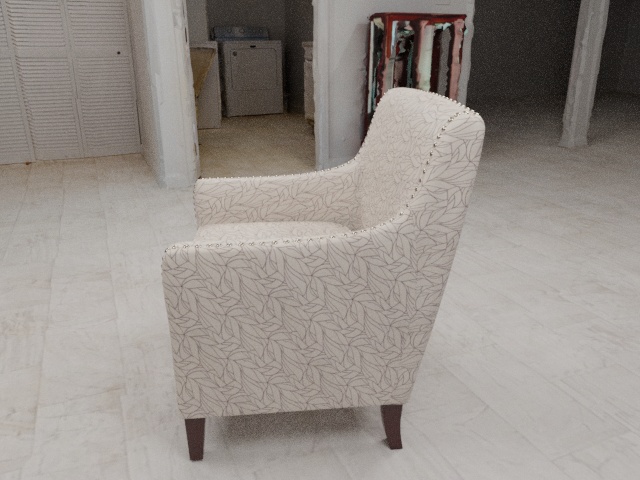} &
          \includegraphics[height=\imh, width=\imw, keepaspectratio]{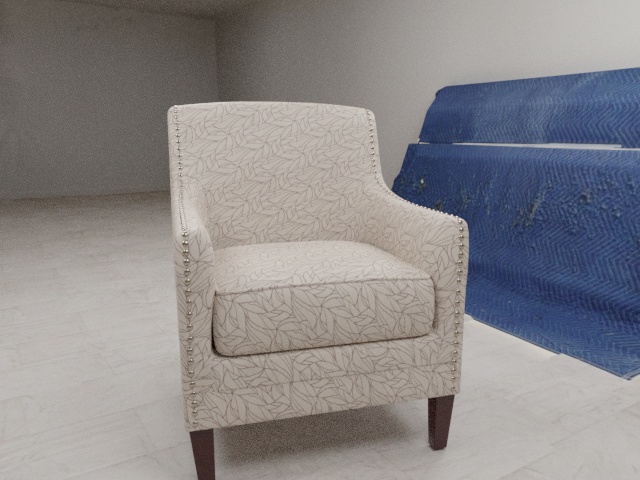} &
           \includegraphics[height=\imh, width=\imw, keepaspectratio]{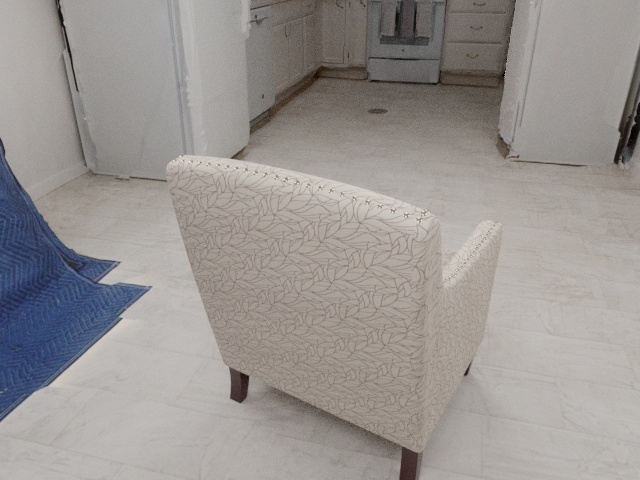} \\

           \includegraphics[height=\imh, width=\imw, keepaspectratio]{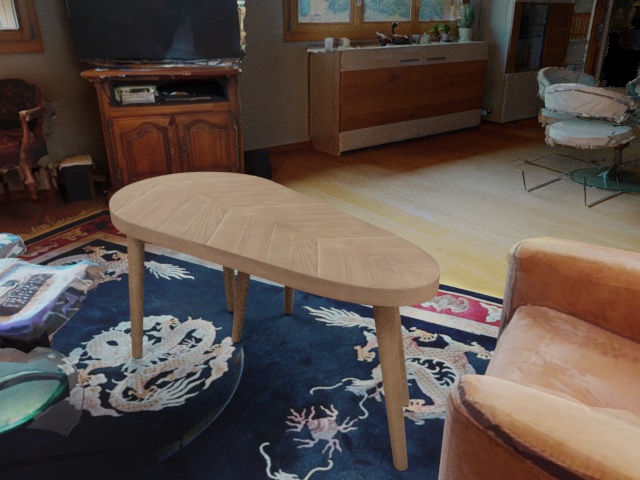} & 
         \includegraphics[height=\imh, width=\imw, keepaspectratio]{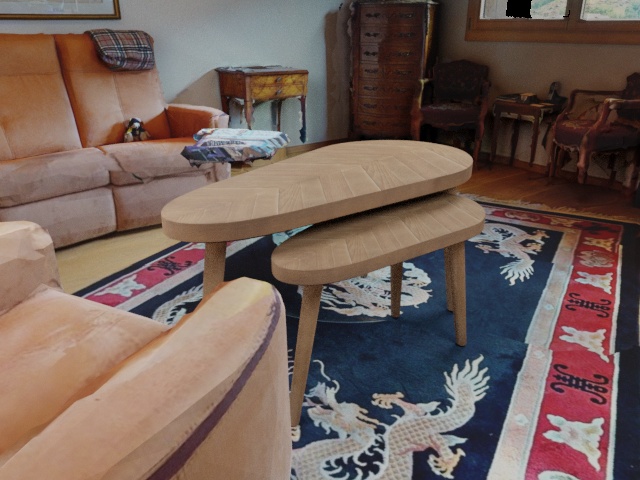} &
          \includegraphics[height=\imh, width=\imw, keepaspectratio]{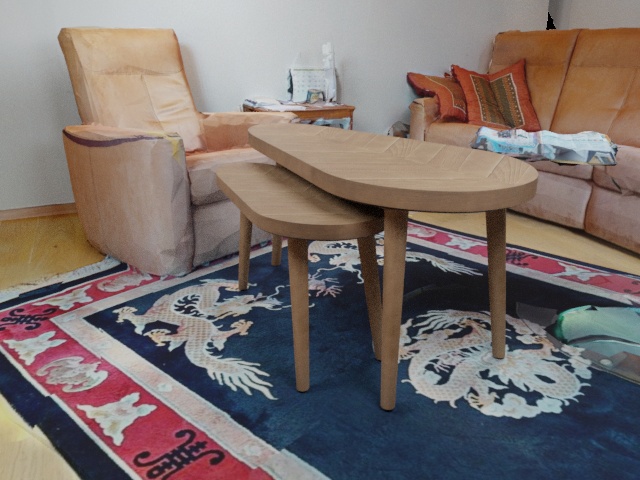} &
           \includegraphics[height=\imh, width=\imw, keepaspectratio]{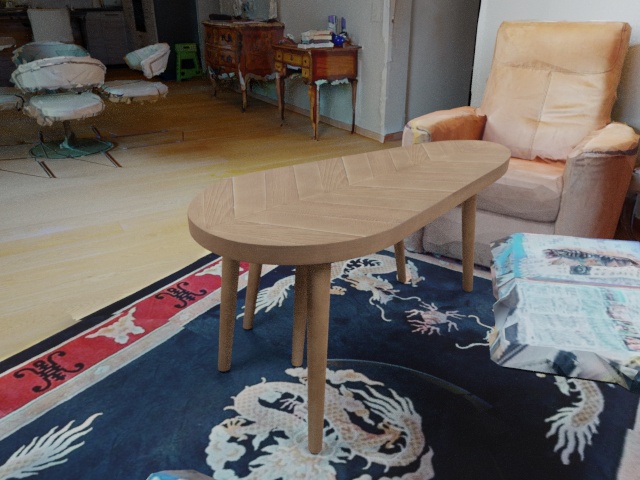} \\

           \includegraphics[height=\imh, width=\imw, keepaspectratio]{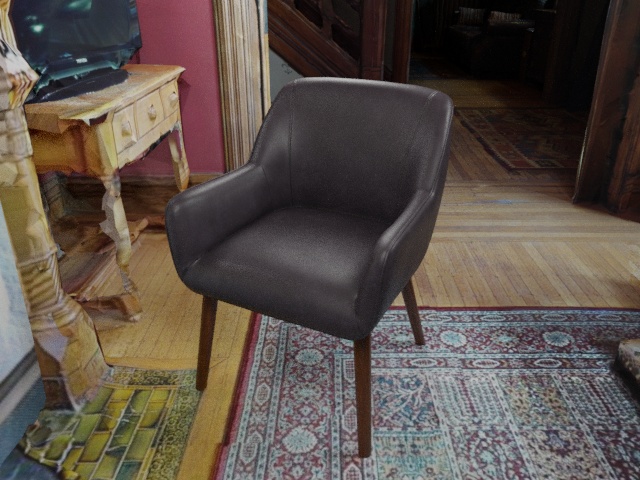} & 
         \includegraphics[height=\imh, width=\imw, keepaspectratio]{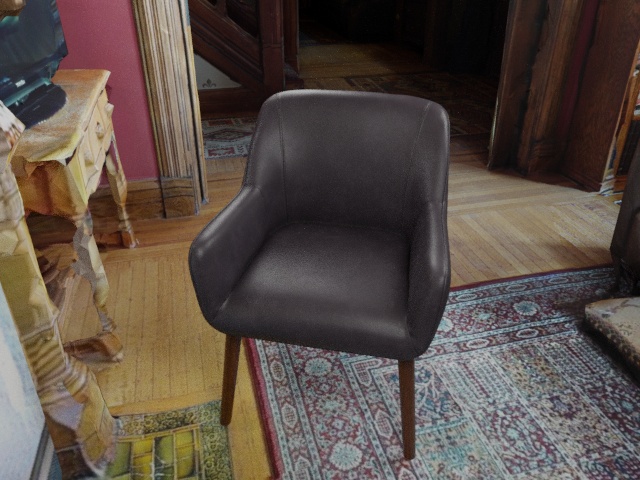} &
          \includegraphics[height=\imh, width=\imw, keepaspectratio]{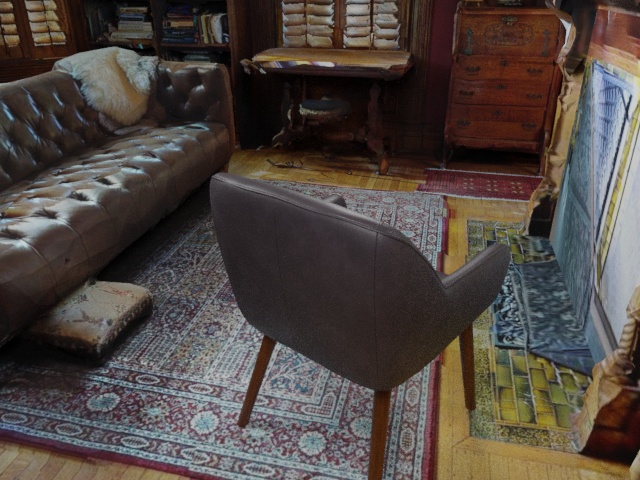} &
           \includegraphics[height=\imh, width=\imw, keepaspectratio]{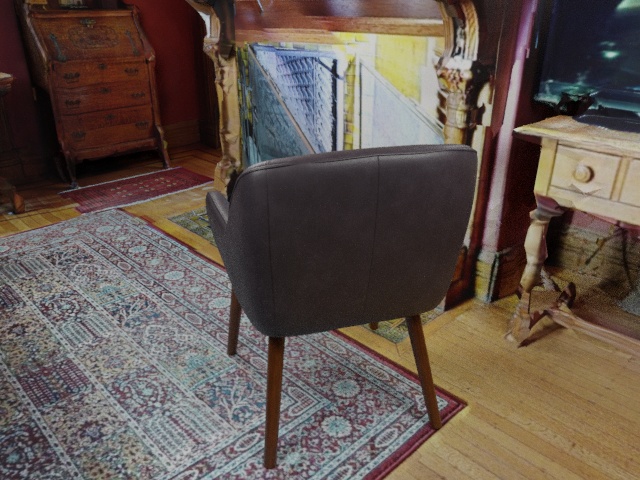} \\

           \includegraphics[height=\imh, width=\imw, keepaspectratio]{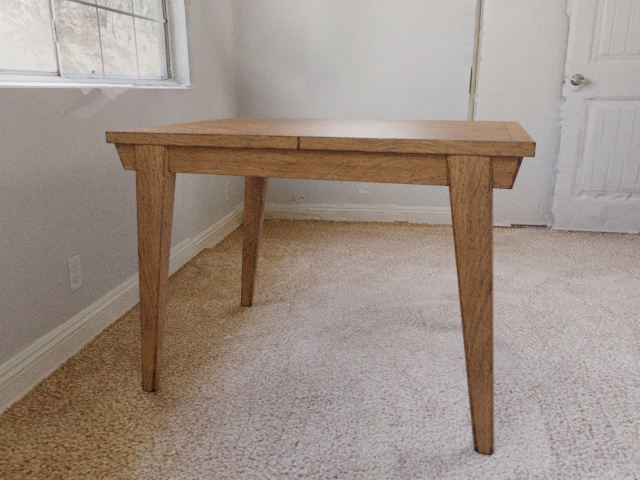} & 
         \includegraphics[height=\imh, width=\imw, keepaspectratio]{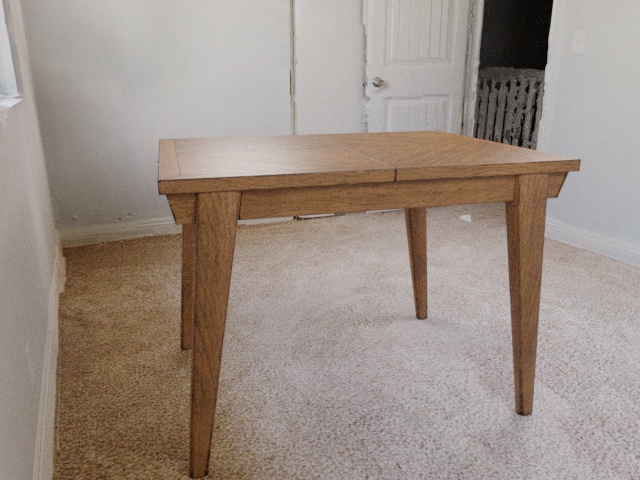} &
          \includegraphics[height=\imh, width=\imw, keepaspectratio]{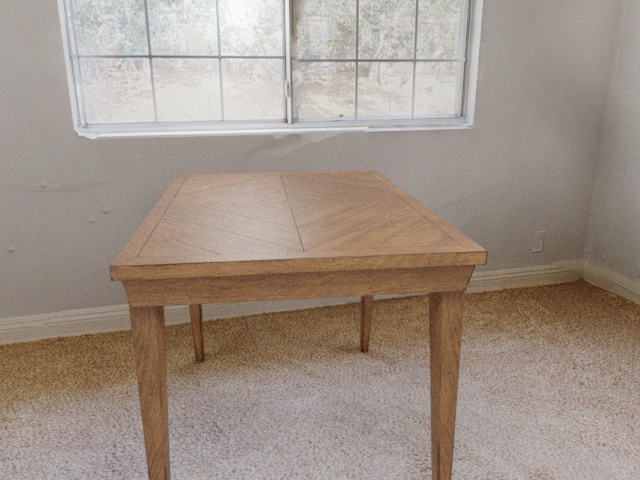} &
           \includegraphics[height=\imh, width=\imw, keepaspectratio]{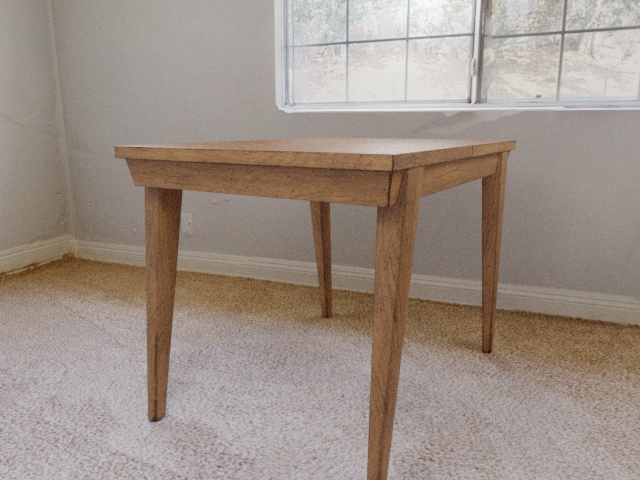} \\

           \includegraphics[height=\imh, width=\imw, keepaspectratio]{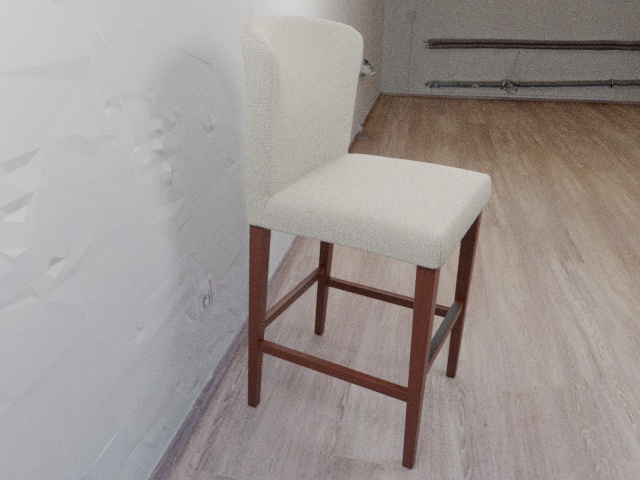} & 
         \includegraphics[height=\imh, width=\imw, keepaspectratio]{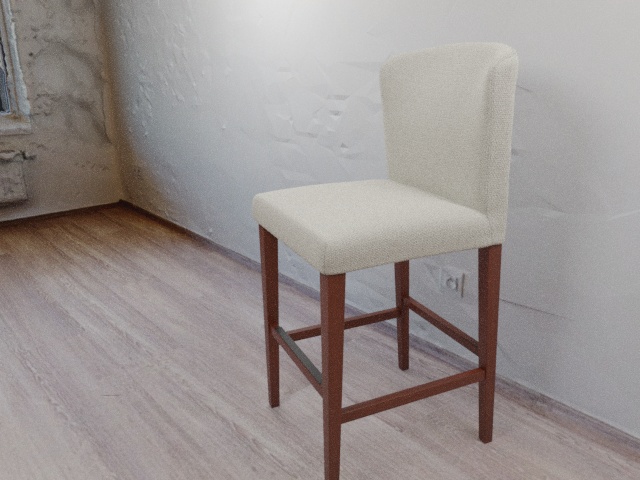} &
          \includegraphics[height=\imh, width=\imw, keepaspectratio]{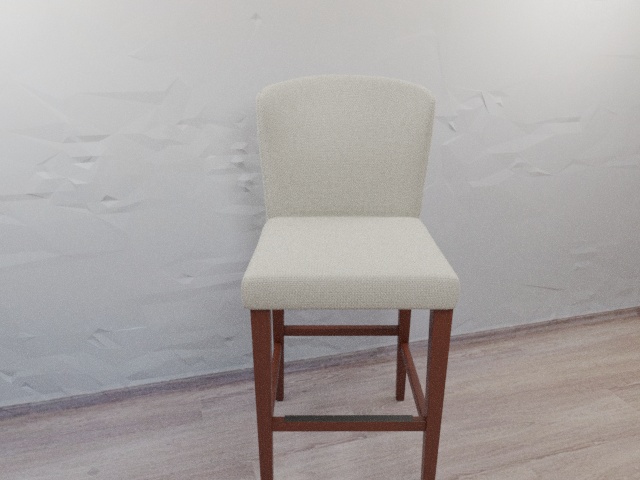} &
           \includegraphics[height=\imh, width=\imw, keepaspectratio]{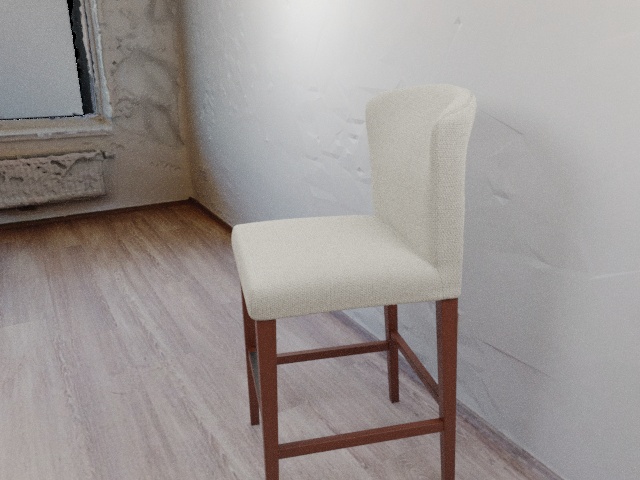} \\
 
           \includegraphics[height=\imh, width=\imw, keepaspectratio]{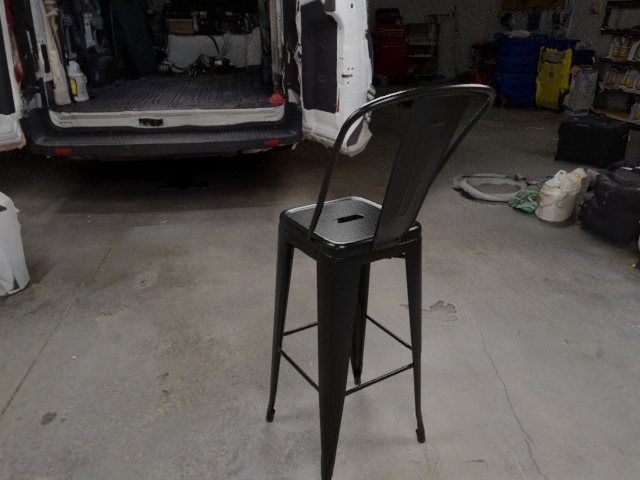} & 
         \includegraphics[height=\imh, width=\imw, keepaspectratio]{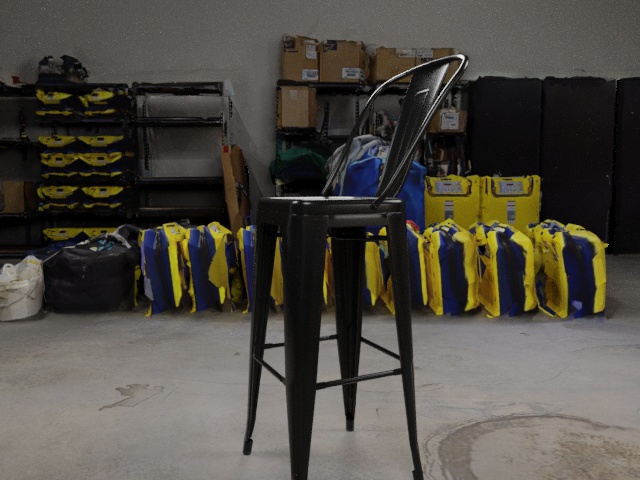} &
          \includegraphics[height=\imh, width=\imw, keepaspectratio]{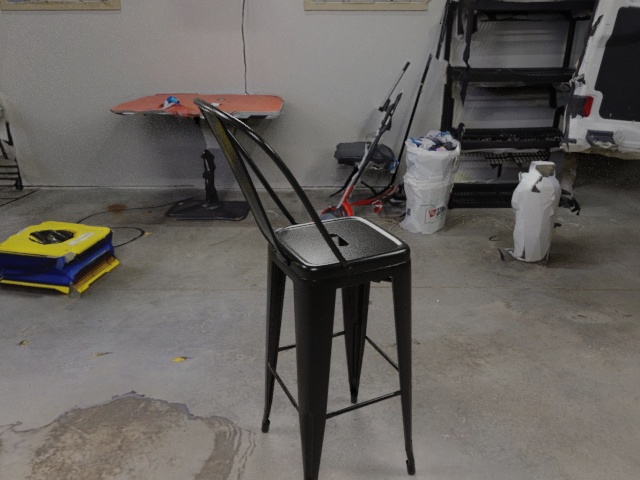} &
           \includegraphics[height=\imh, width=\imw, keepaspectratio]{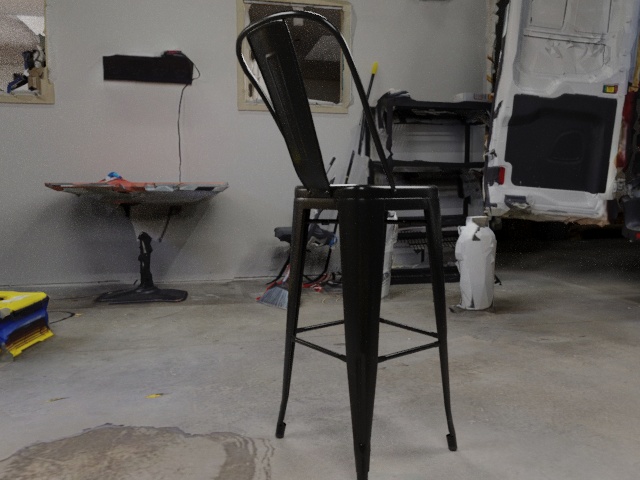} \\

	 \end{tabular}
     \vspace{-5pt}
     \caption{Examples of HM3D-ABO dataset. 3D assets from Amazon-Berkeley Object dataset~\cite{collins2021abo} are placed in scenes from Habitat-Matterport 3D dataset~\cite{ramakrishnan2021habitat} in a physically plausible way. For each object-scene configuration we render multiple images around the object. We show here for each configuration only 4 out of tens views. }
     \label{fig:visualization:matterportabo}\vspace{-1mm}
   \end{figure*}
\endgroup

\subsection{Few-view 3D Object Reconstruction}
Traditional research on 3D object reconstruction mainly focuses on SfM (structure-from-motion) pipelines from dense views, which are difficult to acquire for many applications. Single-view 3D reconstruction~\cite{girdhar2016learning,fan2017point,wang2018pixel2mesh,groueix2018papier,kar2015category,wu2016single,mescheder2019occupancy} receive a lot attention in recent years, but their reconstruction quality is often found to be limited~\cite{what}. Few-view 3D Object Reconstruction recovers the shape of the underlying object given a few RGB observations. Such a setting strikes a balance between capture complexity and reconstruction accuracy, and has been studied in several prior works~\cite{choy20163d,apple,xie2020pix2vox++,disn,zhang2021ners,yang2022fvor}. 

Existing approaches can be categorized into three types. The first type does not use camera pose for each image, such as 3D-R2N2~\cite{choy20163d} and OccNet$^\dagger$, which is an multi-view version implementation of OccNet~\cite{mescheder2019occupancy} proposed by 3D43D~\cite{apple}. The second type assume known camera pose, such as 3D43D~\cite{apple} and FvOR w/ GT Pose~\cite{yang2022fvor}. The other type uses the estimated camera pose, such as FvOR~\cite{yang2022fvor}. In particular, we experimented with the following baselines: 

\noindent\textbf{OccNet$^\dagger$}~\cite{apple,mescheder2019occupancy}. This is a multi-view version of OccNet~\cite{mescheder2019occupancy} by 3D43D~\cite{apple}. It does not require camera pose. 

\noindent\textbf{IDR}~\cite{idr} is an optimization-based algorithm that does not learn a prior from training data. IDR achieves good performance when reconstructing objects with tens of images paired with ground truth object masks. We run IDR~\cite{idr} on each test input for 1000 epochs for best results. 

\noindent\textbf{FvOR w/ GT Pose} is FvOR~\cite{yang2022fvor}'s standalone 3D reconstruction module trained with ground truth camera poses.

\noindent\textbf{FvOR w/o Joint} is FvOR~\cite{yang2022fvor}'s approach without performing joint refinement. 

\noindent\textbf{FvOR} is FvOR~\cite{yang2022fvor}'s approach with joint refinement.

During training, for each scene-object pair, we randomly select 4 views to form an input set. For testing, we select one bag of 4 views from each scene-object pair, yielding $500$ testing examples in total. We compare the IoU metric and the Chamfer-L1 metric~\cite{yang2022fvor,mescheder2019occupancy}. To evaluate the shape quality invariant to similarity transform, we follow the procedure of FvOR~\cite{yang2022fvor} to find the similarity transform that aligns the predicted shape and G.T. shape before evaluation. We also follow FvOR~\cite{yang2022fvor} and compare all methods under three pose initialization settings. The first setting assumes we have access to the G.T. camera pose.  The second setting assumes we have G.T. pose corrupted by different magnitudes of Gaussian noise. The pose perturbation magnitude is controlled by $\sigma\in \big\{0.75\mathrm{e}{-2}, 1.5\mathrm{e}{-2}, 2.25\mathrm{e}{-2}\big\}$ with three values (called Noise@L1, Noise@L2, and Noise@L3 respectively). The last setting uses predicted camera pose by FvOR~\cite{yang2022fvor}'s pose initialization module FvOR-Pose. Note that these three pose initialization settings do not affect OccNet$\dagger$ as it does not utilize the camera pose at all. 

The quantitative results can be found in Table~\ref{table:habitat-shape-refine}. FvOR performs the best under most of the settings. IDR does not give satisfying results, which is probably due to the too-few input views. Pose-free method OccNet$^\dagger$ gives decent results, and the fact that they do not need to estimate camera pose at all makes their method much simpler.

\section{Discussion}
In this report, we introduced a pipeline to create a photo-realistic object-centric dataset, HM3D-ABO. The synthesized images have realistic object and backgrounds compared to dataset based on ShapeNet~\cite{choy20163d,disn} and synthetic scene~\cite{xie2020pix2vox++}. It also comes with high quality 3D models which are absent from several real-captured object-centric datasets~\cite{Choi2016,Reizenstein_2021_ICCV, ahmadyan2021objectron,pix3d}.  Such a dataset can be used for multiple purposes, for example, evaluating algorithms for few-view 3D reconstruction,  novel-view synthesis, and camera pose estimation.  \looseness=-1

While we argue the HM3D-ABO dataset is a more realistic dataset than the previous synthetic object-centric datasets, we also note several limitations. First, the object assets are limited in terms of category. In the current version, most of the objects are chairs and tables, which are the major categories in ABO~\cite{collins2021abo} dataset. With the increased object-level assets we could also expand our HM3D-ABO dataset. Secondly, the current object placement scheme is limited. We only place the single object on an empty region of the walkable surface, which does not include some more challenging scenarios that could also happen in practice, such as reconstructing a teddy bear sitting on the sofa. \looseness=-1

\FloatBarrier
%\clearpage
{\small
\bibliographystyle{ieee_fullname}
\bibliography{refs}
}

% \newpage

\end{document}